 \documentclass[final,1p,times]{elsarticle}

\usepackage{amssymb}
\usepackage{amsmath,amssymb}

\usepackage{xtab}
\usepackage{multirow}
\usepackage{booktabs}
\usepackage{lscape}

\journal{Applied Soft Computing}

\begin{document}

\begin{frontmatter}



\title{Immuno-inspired robotic applications: a review}


\author[af1,af2]{Ali Raza}
\address[af1]{Department of Mechanical Engineering, University of Texas at Austin, USA}
\address[af2]{Department of Mechatronics \& Control Engineering, University of Engineering \& Technology, Lahore, Pakistan}
\ead{ali.raza@utexas.edu}

\author[af1]{Benito R.~ Fernandez}
\ead{benito@mail.utexas.edu}

\begin{abstract}
Artificial immune systems primarily mimic the adaptive nature of biological immune functions. Their ability to adapt to varying pathogens makes such systems a suitable choice for various robotic applications. Generally, AIS-based robotic applications map local instantaneous sensory information into either an antigen or a co-stimulatory signal, according to the choice of representation schema. Algorithms then use relevant immune functions to output either evolved antibodies or maturity of dendritic cells, in terms of actuation signals. It is observed that researchers, in an attempt to solve the problem in hand, do not try to replicate the biological immunity but select necessary immune functions instead, resulting in an ad-hoc manner these applications are reported. Authors, therefore, present a comprehensive review of immuno-inspired robotic applications in an attempt to categorize them according to underlying immune definitions. Implementation details are tabulated in terms of corresponding mathematical expressions and their representation schema that include binary, real or hybrid data. Limitations of reported applications are also identified in light of modern immunological interpretations. As a result of this study, authors suggest a renewed focus on innate immunity and also emphasize that immunological representations should benefit from robot embodiment and must be extended to include modern trends.
\end{abstract}

\begin{keyword}

Artificial immune systems \sep Mobile robots \sep Idiotypic network \sep Clonal selection \sep Danger theory
\end{keyword}

\end{frontmatter}


\section{Introduction}
\label{}

The \emph{biological immune system (BIS)}, consisting of a large number of cells and molecules, works to protect its host from invading infectious agents, known as pathogens. Immune responses to invading pathogens are triggered by the recognition of antigen. \emph{Adaptive} part of BIS handles the invading bacterium or viruses by adapting to varying pathogens, in an antigen-specific manner. Whereas, the \emph{innate} immune system handles common bacteria in a non-antigen-specific manner. Immunologists are still trying to fully interpret the working of BIS, which is based on a wide range of cells and molecules and their interconnections. Resultantly, a number of models have been presented over the years to describe its working as a whole along with the functions of each cell-type. It is because of the fact that BIS exhibits intelligence in terms of self-organization, learning, adaptation, recognition, robustness and scalability\cite{Dasgupta2009}, researchers are trying to implement its computational interpretations in various applications. Artificial immune system (AIS) is, therefore, defined as an adaptive system that is inspired by theoretical immunology and observed immune models, which are applied to problem solving \cite{Castro2002, Hart2008}.

However, intelligence demands embodiment to exhibit itself, just as humans use bodies to show their brains. Robotic systems are, therefore, a common choice among researchers to exhibit intelligent nature of their algorithms. Robotic systems can be wide ranging but in most of artificially intelligent systems, mobile robots are used to navigate through different environments and scenarios. Robot navigation can be classified as either using global path planning or local reactive approach \cite{Ghallab2004}. Global path planning needs prior knowledge of the environment in which robot has to navigate. It means that shape, location and orientation of walls, obstacles, food items and/or targets is required to be known. However, most real world problems are inherently unstructured and thus all the knowledge may not be known ahead of time. On the other hand, local reactive navigation approach uses the local instantaneous information from sensors to help navigate the robot. It gives directives to handle the local situation in terms of steering directions, thus eliminating the need of a-priori knowledge of environment. However, this approach does not necessarily guarantees the solution nor its optimality.

A robot, in a typical navigation experiment, has to arbitrate different behaviors of wander, obstacle avoidance, target seeking, etc. Brook's subsumption architecture \cite{Brooks1986} defines a behavior based reactive approach to control mobile robotic systems allowing intelligence to emerge from behavior arbitration. Like other approaches of neural networks and reinforcement learning, AIS can also be used in conjunction with this architecture to design effective sense-act algorithms. A successful AIS-based robotic system should, therefore, be able to perform behavior arbitration on the basis of underlying immune functions. These immune functions can be based on one of many computational interpretations of BIS, ranging from earlier clonal selection to most recent danger model.

In context of navigating robot(s) in unstructured environments and making them exhibit multiple behaviors, AIS presents a biologically inspired framework that can solve such problems because BIS can adapt to handle unknown pathogens. A number of robotic applications has been published over the years albeit questions on validity of older immune models (e.g. idiotypic network). Recent BIS definitions (e.g. danger theory) are also required to be looked into. Current trends in mobile robotics are also more inclined towards heterogeneity and consequent problem of coordination/cooperation among robots of various capabilities \cite{Baldassarre2003,Tuci2008,Wang2008}, over and above the classic problem of navigation in unknown environments. Other problems include, but are not limited to, conflict resolution in multiple behaviors \cite{Pallottino2007,Powers2010}, probabilistic robotics \cite{Thrun2005} and behavior evolution \cite{Konig2009}. Such problems are testing the extents of control/navigation algorithms. It is, therefore, important to review immuno inspired robotic applications in light of emergent aspects of BIS and newer trends in robotics.

\begin{figure}[ht]
\begin{center}
\includegraphics[width=3.0in, height=2.4in]{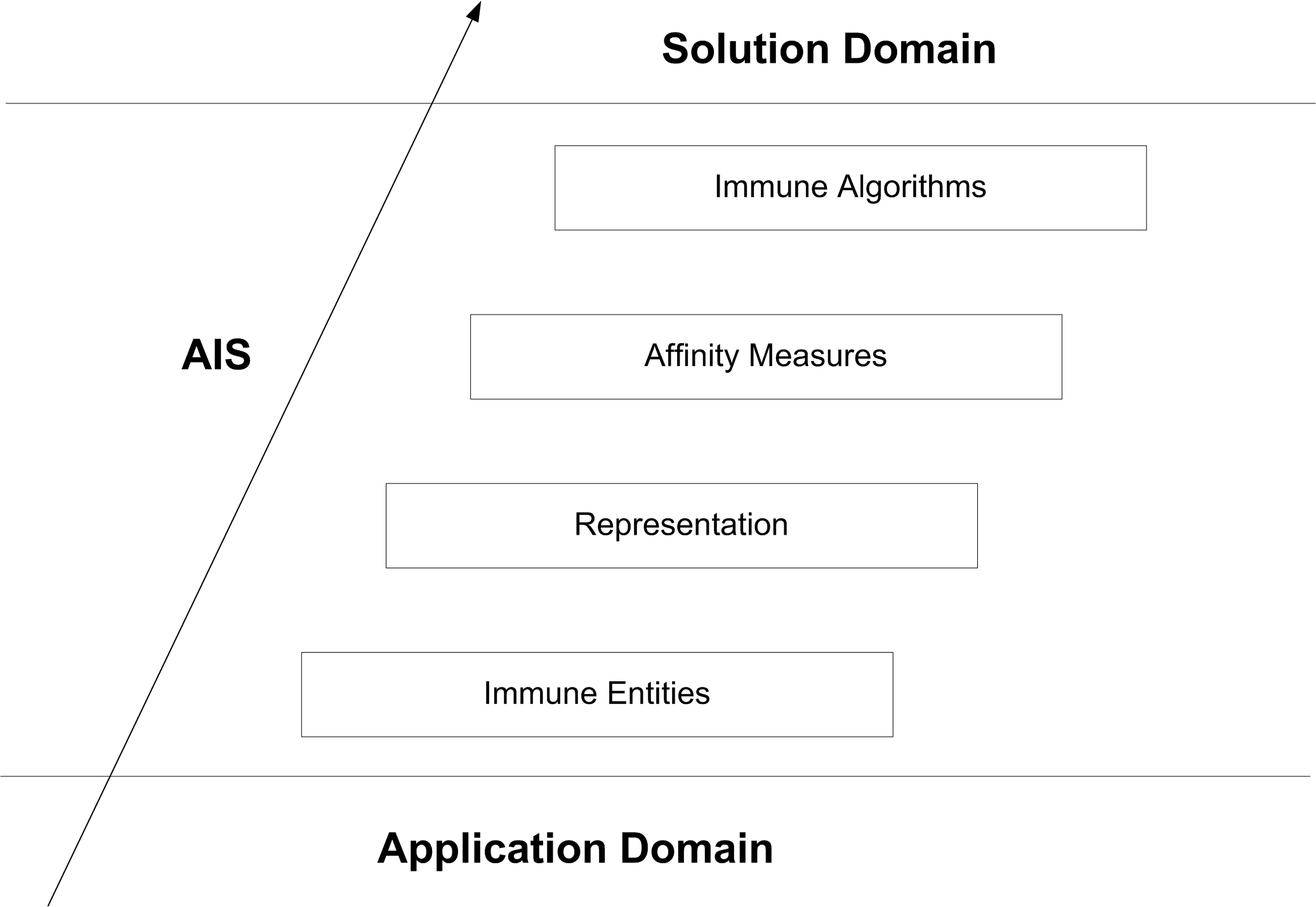}
\end{center}
\caption{AIS-based framework for problem solving, adapted from Castro \cite{Castro2002}.}
\label{figure_CastroFW}
\end{figure}

Castro \cite{Castro2002} defined a layered framework in which an AIS undergoes the processes of representation, affinity measures and immune algorithms to solve a problem in application domain, as shown in fig. \ref{figure_CastroFW}. Robotic applications, using AIS as a core algorithm, generally use the same framework. This review uses Castro's framework as a selection guideline for the reported literature by tabulating details in a structured format. However, some applications do not follow the framework by either avoiding the underlying details or using AIS only as a metaphorical explanation of their algorithms; those are resultantly skipped in this review.

This report is also an effort to analyze the literature in terms of underlying AIS definitions, representation schema, application type and corresponding results. Report also discusses the implementation issues in such applications. Subsequent section detail out computational interpretation of various AIS models along with their corresponding general algorithms. Section 3 categorizes the reported robotic applications in terms of their AIS definitions and underlying mathematical details. Section 4 presents a discussion to identify the established results as well as the areas that require further investigation.

\section{AIS Definitions:}

There are a number of explanations of BIS and a lot of research is still being done by immunologists all over the globe. The goal of immuno inspired computational research, on the other hand, is to translate such definitions to solve problems. Clonal selection theory \cite{Burnet1959} is the oldest definition that interprets the working of B cells in a BIS. It was augmented by negative selection to describe the phenomenon when BIS chooses not to respond, to give rise to self-nonself theory \cite{Bretscher1970}. Immune network theory \cite{Jerne1974}, most commonly used in robotic applications, defined the working of antibody network that enables antibodies to recognize each other. Recent major development in immunology is the inclusion of danger theory \cite{Matzinger2002} to construct a three signal approach to handle invading pathogens in dangerous/stressed situations. Following subsections briefly define the AIS models that are used in robotic applications.

\subsection{Clonal Selection:}

\begin{figure}[ht]
\begin{center}
\includegraphics[width=3.2in, height=2.6in]{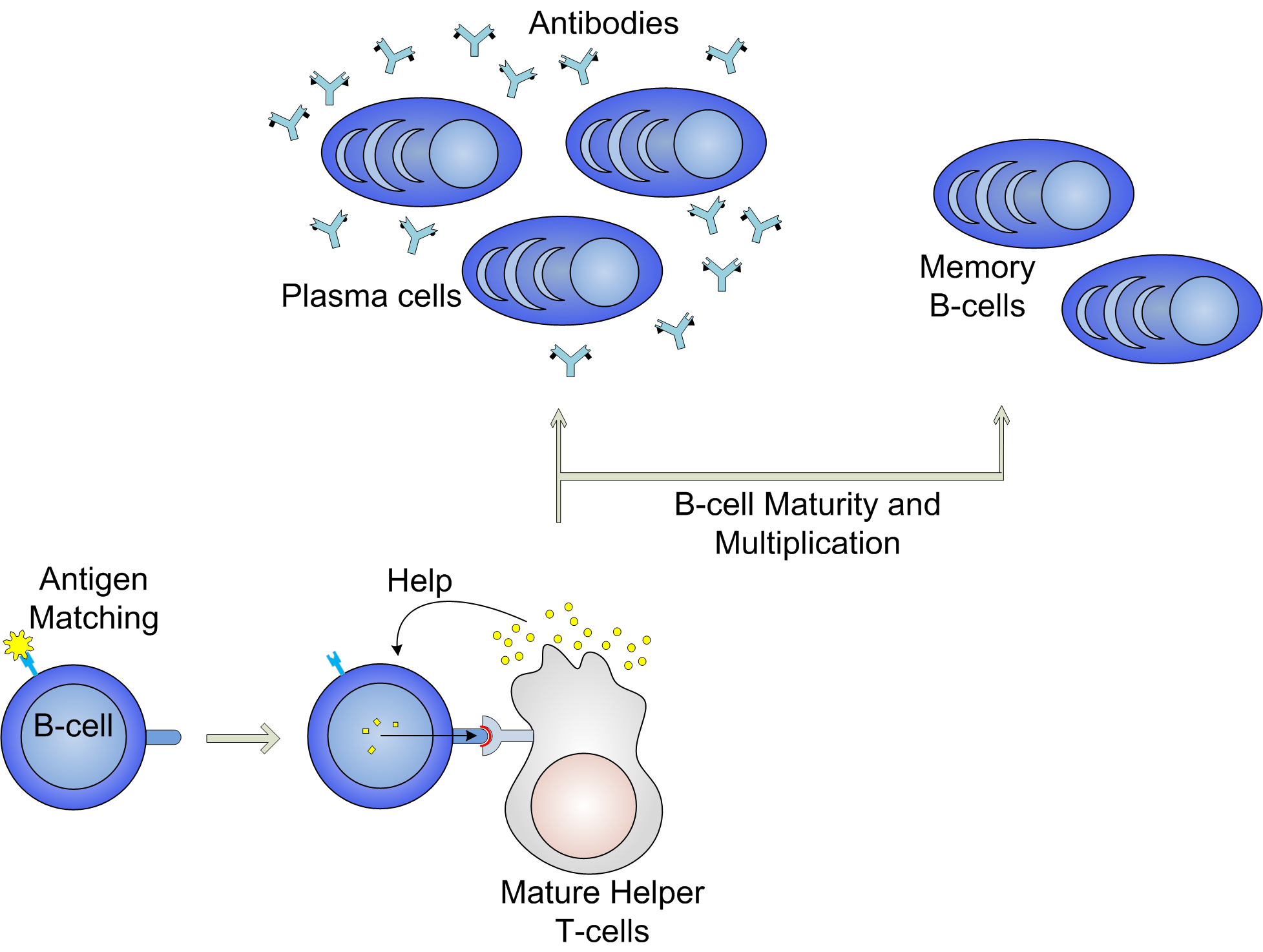}
\end{center}
\caption{Working of B-cells in a biological immune system}
\label{figure_Bcells}
\end{figure}

Clonal selection (CS) theory is one popular explanation of how B and T lymphocytes improve their response to presented-antigens in order to acquire immunity through affinity maturation. Selection is inspired by the antigen-antibody-affinity. It states that B-cells divide when an affinity is present between stimulating antigen's epitope and B-cell receptors. These cells then mature into plasma cells and secrete antibodies. Antibodies with higher affinities are then reproduced through somatic hypermutation of B-cells. Paratopes on antibodies and epitopes on antigens work as key-lock mechanism (complement cascade) to help other cells to eliminate antigens. Immune system retains some matching B-cells as memory cells. Moreover, it adapts by building up concentrations of B-cells as well as maintains a diversity in mutating these cells in the bone marrow \cite{Dasgupta2006}. Figure \ref{figure_Bcells}, constructed from various sources including \cite{2007,Janeway2001,Matzinger2002}, provides a description on the working of B-cells once antigen is presented to it.

\subsubsection{Computational Interpretation of Clonal Selection}

It is important to mathematically interpret affinity between antigen and antibody. Selection, ordering and subsequent reselection of antibodies or mutated antibodies is solely done on the basis of affinity scores. It acts similar to fitness function in genetic algorithms (GA). Affinities $A_{f_i}$ are application specific but generally can take the following form.
\begin{equation}\label{eq_affinity}
    A_{f_i} = \frac{A}{\sum_{i=1}^{N}\left(d_i + c \beta_i\right)}
\end{equation}
Where $d_i$ is distance between presented antigen and selected antibody and $\beta_i$ can be defined in terms of available auxiliary data. Commonly, the distance is translated in terms of euclidean or hamming distances based on real or binary representations, respectively. Clonal selection is adaptive and works on the principle of antibody evolution through somatic hypermutation. The results of affinity computations, using Eq. \ref{eq_affinity}, are sorted in ascending order which is followed by reselection on basis of best-population-size and subsequent maturation using Eq. (\ref{eq_Mu}). Each antibody in selected-and-ordered-best-population is then cloned as described in Eq. \ref{eq_Clone}. These clones are projected within the solution bounds. Affinities are computed again and resulting best clones are selected. Selected best clones then replace the antibodies in initial antibody matrix.

\begin{eqnarray}
\mu_i &=& K_1 e^{-K_2 \cdot A_{f_i}} \label{eq_Mu}\\
C_i &=& A_i + \gamma \left[\mu_i \cdot rand(.)\right] \label{eq_Clone}
\end{eqnarray}

In this expression, $\mu_i$ is antibody maturation rate, $K_1$ is maturation constant and $K_2$ is maturation decay factor. Whereas, $C_i$ and $A_{f_i}$ are clones and affinity of $i$th selected antibody respectively. $\gamma$ is scaling factor for random number generator for the cloning expression. There can be other variants of maturation and cloning expressions (Eq. \ref{eq_Mu} \& \ref{eq_Clone}) on the basis of corresponding representation schema.\\

For a comprehensive computational detail on clonal selection, reader is referred to White and Garrett \cite{White2003}. Garrett \cite{Garrett2003} also presented an alternative representation to combine several B-Cell representations in an attempt to combine the clonal selection and immune network approaches in a generic network.

\subsubsection{General Algorithm of Clonal Selection}
Although there are a number of variants of CS-based algorithms, following table \ref{table_CSAlgo} is one generic algorithm. It starts with the initialization of antibody parameters. An antibody matrix is specified at this stage by including the initial solution candidate as well as randomly generated antibodies. Best antibodies are then selected on the basis of affinity evaluation. Affinity function uses the information of antibody \& antigen and calculates the affinity on the basis of selected function e.g. boolean operators, euclidian /hamming distance or a user defined function. A number of string matching rules are listed by Dasgupta \cite{Dasgupta2009} for interested readers. The premise of these definitions is to use a function that best incorporates the key-lock mechanism of antigen-antibody interaction. The rest of algorithm follows the procedure described in previous subsection.
\begin{table}[h!]
\caption{General Clonal Selection Algorithm}
\begin{center}
\label{table_CSAlgo}
\scriptsize
\begin{tabular}{p{8cm}}
\hline \hline
Algorithm\\
\hline
\item {\bf Input:} Antigen: $A_g$
\item {\bf Output:} Evolved antibodies $A_b$ (Memory)
\begin{enumerate}
  \item {\bf Initialize:} Random population of $A_b$
  \item {\bf While} $\neg$ goal {\bf do}
  \begin{itemize}
    \item {\bf for} all Antigens $A_g$ {\bf do}
  \begin{itemize}
    \item  Affinity computation: for each $A_b$
    \item  Selection: according to affinities
    \item  Cloning: proportional to respective affinity
    \item  Maturation: mutate clones, inversely proportional to corresponding affinities
    \item  Reselection: according to affinities
    \item  Update Memory
    \item  Metadynamics
  \end{itemize}
    \item{\bf end}
    \item{\bf repeat}
  \end{itemize}
  \item {\bf end}
\end{enumerate}\\
\hline
\normalsize
\end{tabular}
\end{center}
\end{table}

\subsection{Immune Network:}
\begin{figure}[ht]
\begin{center}
\includegraphics[width=3.2in, height=1.6in]{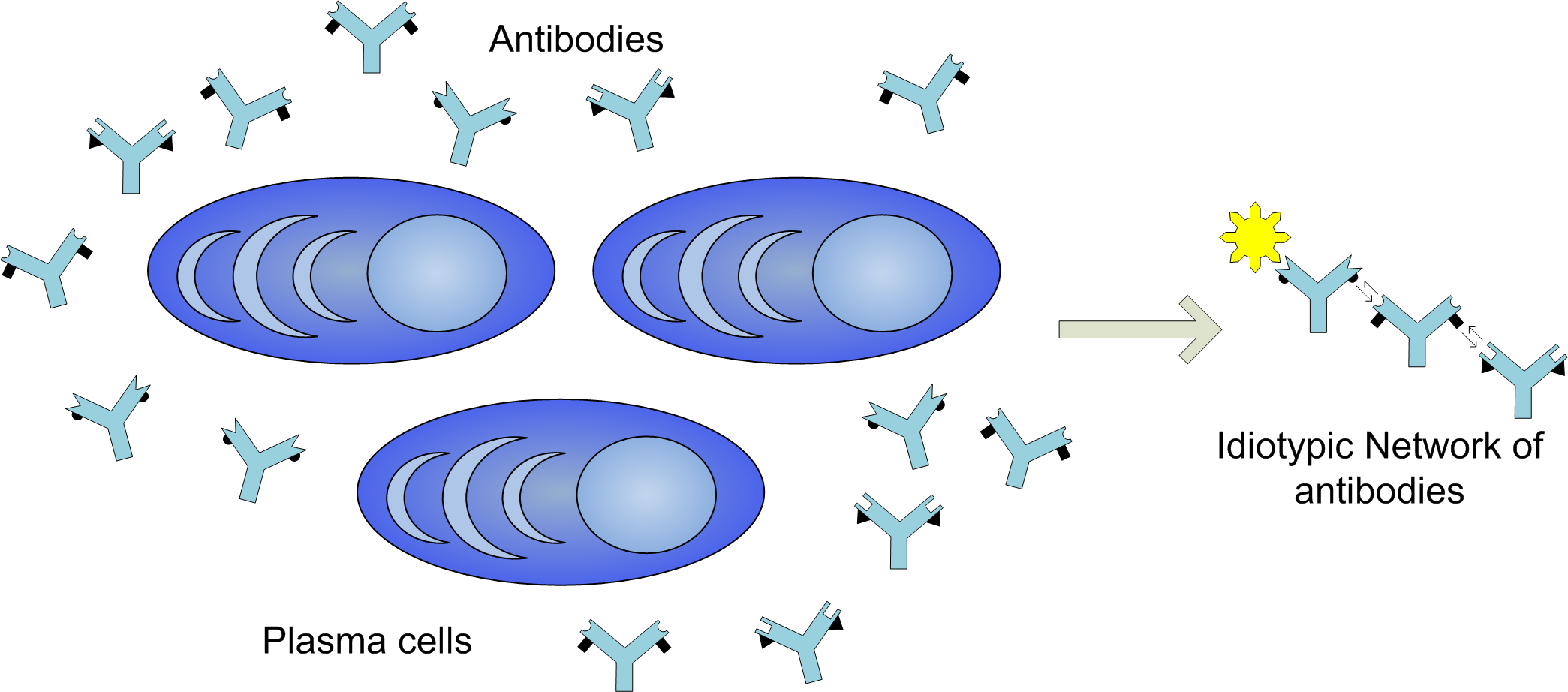}
\end{center}
\caption{Working of idiotypic stimulations and suppressions in an Immune Network}
\label{figure_IN}
\end{figure}

Clonal selection theory does not explain the working BIS in absence of invading pathogens or suppression of certain immune functions. Jerne's idiotypic-network theory \cite{Jerne1974}, also known as immune network (IN), proposes the possible explanation. It suggests that an antibody possesses a unique idiotope, similar to epitope, so that other antibodies can recognize it. The group of antibodies that share common idiotope belongs to one idiotype. This theory also states that once an antibody's idiotope is recognized by paratopes of other antibodies, it is suppressed. Consequently, antibody concentration is reduced. Similarly, once an antibody's paratope recognizes idiotopes of other antibodies or epitopes of antigens, it is stimulated. Antibody concentration is increased as a result of this stimulation. In other words, this theory tries to explain the communication between antibodies via collective dynamic network of stimulative and suppressive interactions, suggesting a continuous communication even in absence of antigens. This is in contrast to the $\it{antibody-antigen-only}$ interactions of clonal selection theory. It is because of this notion that cells within an immune system can recognize each other, in addition to recognizing antigens, this theory is applied on a number of different applications ranging from internet security to mobile robotic systems. Figure \ref{figure_IN} illustrates the network of antibodies using idiotypic connections.

\subsubsection{Computational Interpretation of Immune Network}

Jerne's idiotypic network theory \cite{Jerne1974} was translated into a computational model by Farmer et al. \cite{Farmer1986}. Following differential equation was proposed by Farmer et al. for antibody concentration $\it{A_i}$ with respect to all the stimulatory and suppressive effects as well as the natural death rate. \\

\begin{eqnarray}
    \dot{A_i}  =  \left[\alpha_a\sum_{j=1}^{N_a} m_{ij} a_j
                     - \alpha_s\sum_{j=1}^{N_a} m_{ji} a_j
                     + \sum_{k=1}^{N_g} n_{ik} y_k
                     - \lambda_i\right]a_i
                     \label{eq_Farmer1}\\
         a_i   =  \sigma(A_i) = \frac{1}{\tau_i+exp(-A_i)}, ~~~\forall ~i=1,\dots,N_a. \label{eq_Farmer2}
 \end{eqnarray}

It is defined for $N_a$ antibodies and $N_g$ antigens. First sum in Eq.~(\ref{eq_Farmer1}) represents the stimulation of antibody $\it{A_i}$ in response to the other antibodies $\it{A_j}$ (idiotope-paratope connection). It is termed as $\it{stimulus_1}$ in subsequent sections. Second sum represents $\it{suppression}$ of antibody $\it{A_i}$ in response to all other antibodies (paratope-idiotope connection). Third sum models the stimulation of antibody $\it{A_i}$ in response to all antigens (paratope-epitope connection) and is termed as $\it{stimulus_2}$. Last term in the expression shows antibody death rate. The resulting antibody concentration rate depends on the collisions between antibody $A_i$ and antibody $A_j$ that is proportional to $a_ia_j$. Eq.~(\ref{eq_Farmer2}) is a squashing function that controls the size of $a_i$.

Jerne's network theory has some critics as well. It is argued that a very large antibody population limits the network size in idiotypic suppressions \cite{Langman1986}. Similarly, the network structure in terms of its symmetry has its own share of arguments \cite{DeBoer1989}.

\subsubsection{General Algorithm of Immune Network}
There can be a number of variants of IN based algorithms. Following table \ref{table_INAlgo} is the core algorithm that generally is common to all variants.

\begin{table}[h!]
\caption{General Immune Network Algorithm}
\begin{center}
\label{table_INAlgo}
\scriptsize
\begin{tabular}{p{8cm}}
\hline \hline
Algorithm\\
\hline
\item {\bf Input:}
\begin{itemize}
\item Definitions: Antigen, Antibody
\item Antigen: $A_g$
\end{itemize}

\item {\bf Output:} A network of Antibodies: $A_b$

\begin{enumerate}
  \item {\bf While} $\neg$ goal {\bf do}
  \begin{itemize}
    \item Collect Antigen
    \item Affinity: between Antigen and Antibody
    \item Network: Stimulations and Suppressions (Some or all)
    \begin{itemize}
    \item Stimulus 1: between antibodies
    \item Stimulus 2: between antigen and antibodies
    \item Suppression: between antibodies
    \end{itemize}
    \item Antibody Network Dynamics
    \item Cloning
    \item Metadynamics
    \item Return Network
 \end{itemize}
  \item {\bf repeat}
  \item {\bf end}
\end{enumerate}\\
\hline
\normalsize
\end{tabular}
\end{center}
\end{table}

\subsection{Danger Theory:}

\begin{figure}[ht]
\begin{center}
\includegraphics[width=3.2in, height=1.1in]{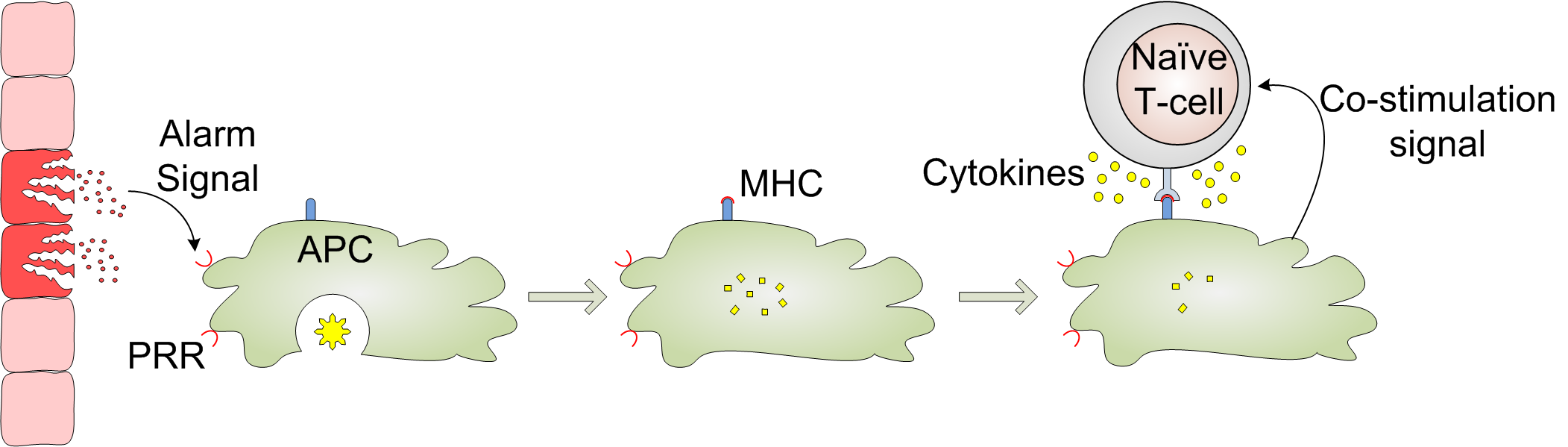}
\end{center}
\caption{Initiation of biological immunity according to danger theory}
\label{figure_DT}
\end{figure}

Newer definitions of self-nonself models and danger theory are extensions of earlier clonal selection theory of antibody production of activated B-Cells. Matzinger \cite{Matzinger2002}, in her danger theory (DT), explains how immune response is initiated. It presents a three signal model to explain the working of biological immunity in stressed situations. It describes that a co-stimulatory signal from dendritic cells activates the T-helper cells. These dendritic cells of the immune system, also termed as antigen presenting cells (APCs), are themselves activated by danger signals emitted by the injured/stressed cells. Once activated, they provide a co-stimulatory signal to exhibit innate/adaptive immune response. Furthermore, dendritic cells can be immature, semi-mature and mature. Immature dendritic cells collect antigens along with safe and danger signals from its local environment like pathogen associated molecular patterns signals (PAMPS) and inflammatory cytokines. If environment is safe, the dendritic cell becomes semi-mature and upon presenting antigen to T-cells it causes the T-cell-tolerance. On the other hand, if environment is dangerous, it becomes mature and causes T-cell-reactivity. Figure \ref{figure_DT} illustrates the maturity of APCs in case of a danger signal that results in co-stimulation of T-cells.

\subsubsection{Computational Interpretation of Danger Theory}
Computational interpretations and consequent algorithms, inspired by danger theory, are still in their infancy. There are two interpretations following introductory work of Aickelin et al. \cite{Aickelin2003}; one is dendritic cell algorithm (DCA) by Greensmith et al. \cite{Greensmith2006} while the other is toll like receptor algorithm (TLR) by Twycross \cite{Twycross2007}. Both use different aspects of danger theory. DCA introduces the underlying translations of PAMPS, safe and danger signals resulting from maturity of dendritic cells with the help of co-stimulatory molecule (CSM). These signals are buffered as well as the antigen. DCA, on the basis of dendritic cell maturity and migration threshold, sets the cell context. Equation \ref{eq_Julie}, in one possible configuration of output $(O_p)$, thus contextualizes the environment which then arbitrates the immune responses. Also see table \ref{table_DCAAlgo} for the algorithm.
\begin{equation}
O_p = \beta\left[W_P\sum_{i=1}^{3}P_i+W_D\sum_{i=1}^{3}D_i+W_S\sum_{i=1}^{3}S_i\right],~~~\forall ~p\label{eq_Julie}
\end{equation}

\subsubsection{General Algorithm of Danger Theory}
At the heart of danger model of immunology is the antigen presenting cells (APC)/dendritic cells. The core of DCA is presented here in table \ref{table_DCAAlgo}. It is a population based algorithm that uses the concept of immune function initialization in a three signal model.

\begin{table}[h!]
\caption{General Dendritic Cell Algorithm, from \cite{Greensmith2006}}
\begin{center}
\label{table_DCAAlgo}
\scriptsize
\begin{tabular}{p{8cm}}
\hline \hline
Algorithm\\
\hline
\item {\bf Input:} Sorted antigen and signals (PAMP,DS,SS)
\item {\bf Output:} Antigen and their context (0/1)

\begin{enumerate}
    \item {\bf Initialize:} DC
    \item {\bf While} CSM output signal $<$ migration threshold {\bf do}
  \begin{itemize}
    \item Collect Antigen
	\item get signals
	\item calculate interim output signals
	\item update cumulative output signals
    \item update cell location to lymph-node
  \item {\bf if} semi-mature output $>$ mature output {\bf then}
    \begin{itemize}
	   \item cell context is assigned as 0
    \end{itemize}
  \item {\bf else}
    \begin{itemize}
	   \item cell context is assigned as 1
    \end{itemize}
  \item {\bf end}
  \item arbitrate the behavior
  \item Invoke innate / adaptive immune response
  \item kill cell
  \item replace cell in population
  \end{itemize}
  \item {\bf end}
\end{enumerate}\\
\hline
\normalsize
\end{tabular}
\end{center}
\end{table}

\section{AIS-based Robotic Applications:}
\begin{figure}[ht]
\begin{center}
\includegraphics[width=3.2in, height=3.0in]{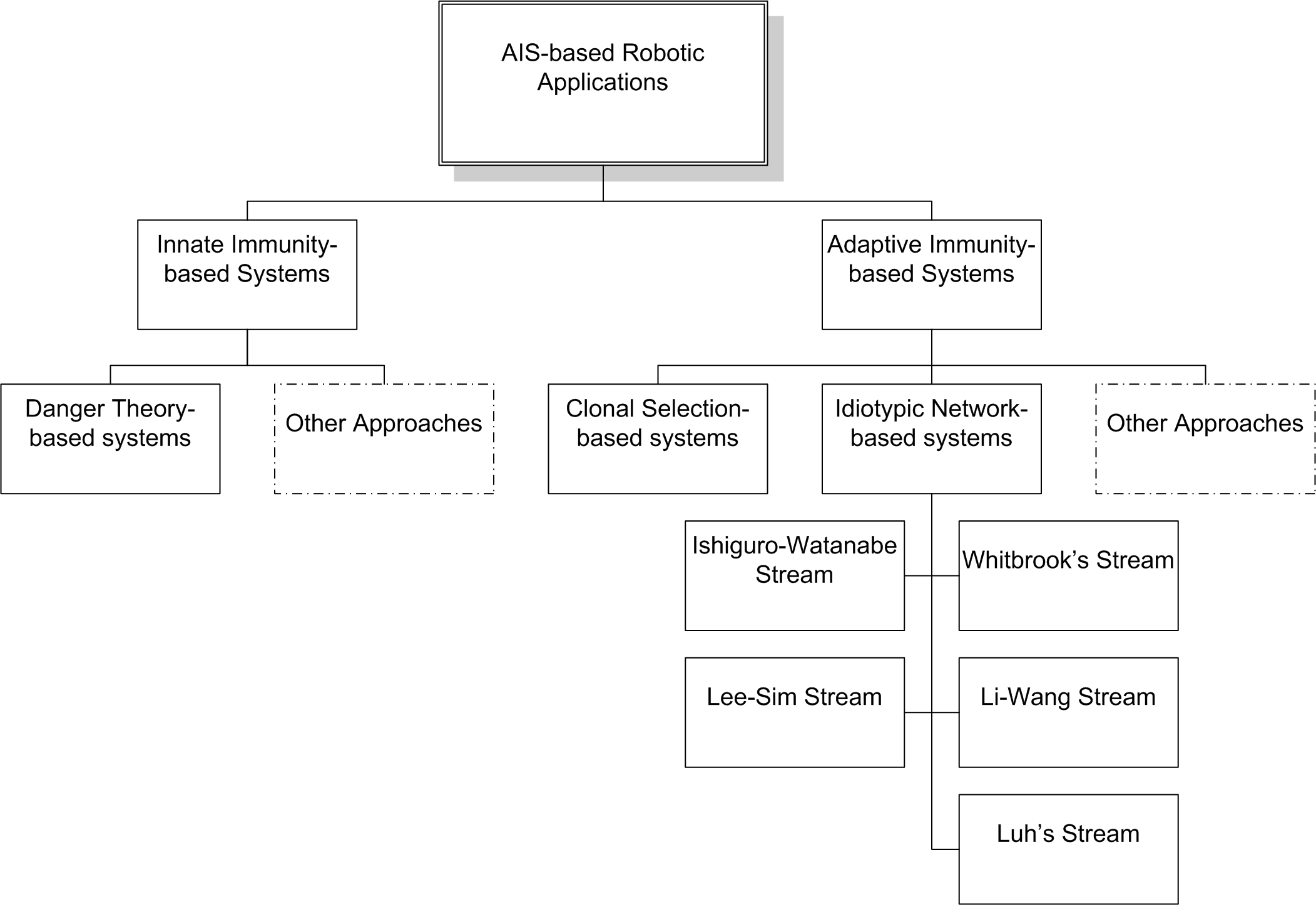}
\end{center}
\caption{Genealogical Chart of AIS-based Robotic Applications}
\label{figure_Genealogy}
\end{figure}

In literature, AIS-based robotic applications tend to simulate robot control around small, artificial environments, generally addressing the problems of behavior arbitration and autonomous navigation (e.g. \cite{Luh2008, Lau2007}). These environments are generally programmed as fixed and depicted as arenas where robots have to perform. Subsequent discussion categorizes the reported immuno-inspired robotic applications in four categories. First category lists the applications using clonal selection as central algorithm, second details those with immune network, third describes those with danger theory and fourth category details those with hybrid/implicit definitions under \emph{other approaches} in both innate and adaptive immunities. Figure \ref{figure_Genealogy} shows a genealogical chart of these categories according to their parent immune categories; innate or adaptive. The discussion is further augmented with underlying mathematical expressions of different approaches in tables \ref{table_CS_review1}, \ref{table_IN_review1} \& \ref{table_IN_review2}.

\subsection{Robotic Applications using Clonal Selection:}
There are different versions of CS algorithms being used by researchers. The variations are mainly due to definitions of antigen-antibody strings, affinity computations and auxiliary functions. Computational efficiencies are also important but mainly depend on the string lengths and population size of mutated antibodies during the process. Subsequent discussion details salient features of reported literature with major criticism on underlying immunological interpretation and should be read in conjunction with table \ref{table_CS_review1}.

Hu \cite{Hu2008} used a CS-based approach for global path planning of a robot. Antibodies are defined as a set of nodes that represent line segments from starting point to end point. Interestingly, this representation does not use antigen definition. Fitness function (affinity) is defined in terms of euclidean distance $d_i$ and obstacle information $\beta_i$ in the arena. Paths that intersect obstacles affect the fitness function.

Wang \& Hirsbrunner \cite{Wang2003} developed an immune mechanism based evolution algorithm (IMEA) in an off-line robot navigation task, in an attempt to avoid premature convergence during navigation, and compared it with genetic algorithm approach. IMEA combines the concepts of genetic algorithms (crossover and mutation) with those of clonal selection (memory updation and selection). It is noted that this approach uses the concept of vitality $\nu_i$ to compute fitness function and is based on least mean squared error between selected two paths. IMEA does not use the concepts of somatic hypermutation and cloning to evolve the solution but relies on the concepts of GA, instead.

Li et el. \cite{Li2005} presented a CS-inspired approach for concurrent mapping and localization in order to search the space for possible robot maps. This approach does not use the metaphors of antigen or antibody but chooses chromosomes to represent change in distance and orientation as in genetic algorithms. It then uses CS for mutation purpose only. In fitness function, described in table \ref{table_CS_review1}, $w_1$ and $w_2$ are real numbers in range of $(0,1)$ and if $O_{ij}>0.5$ then $\delta_{ij}=1$ else $\delta_{ij}=0$; and if $E_{ij}>0.5$ then $\zeta_{ij}=1$ else $\zeta_{ij}=0$.

Hur, J. \cite{Hur2007} developed a multi robot system, for a bomb disposal task, using CS-based antibody evolution to update a lookup table (memory) that lists solutions corresponding to different states. Diffuser-bot, scanner-bot and inspector-bot maximize the number of bombs diffused, number of bombs found and number of bombs inspected, respectively, in addition to maintaining energy levels through affinity computation. The underlying definitions are given in table \ref{table_CS_review1} in which affinities are computed through assigning weights ($w_i$) to a number of certain soft constraints ($n_i$). Number of clones ($N_c$) is proportional to the affinity of individual antibodies in population $N$. The maturated and reselected antibodies then update the lookup table.

There are some robotic applications that use clonal selection principle as an auxiliary function or as a metaphor only. Chingtham and Nair \cite{Chingtham2009} tried a two robot line tracking system using two concentric tracks. Outer robot uses the innate immune system to help inner robot to avoid sway and inner robot uses CS to adjust its speed on track. Literature does not describe the mathematical representation of underlying principles. Jun et al. \cite{Jun1999} used CS metaphors to augment immune network, only to transfer strategy between individual robots.


\subsection{Robotic Applications using Immune Network:}
Immune network is widely used AIS-based approach in robotics, more than any other model, because it explicitly defines all the interactions between antibodies and antigen and their resulting network. The applications are subdivided in terms of their parent technique. The representation schema of antigen and antibody, affinity computation, antibody selection criterion is presented in table \ref{table_IN_review1} along with the details of experiment. Table \ref{table_IN_review2} provides mathematical expressions for stimulations and suppressions in network along with the details of corresponding auxiliary functions, cloning and metadynamics.

\subsubsection{Ishiguro-Watanabe Stream:}
This subcategory deals with binary representations for antigen and antibody. Resultant network uses hamming distance as primary criterion for affinity computation. Consequent antibodies are then selected on the basis of their respective concentrations. The underlying principle of suppressions and stimulations remain similar with minor differences.

Ishiguro et al. \cite{Ishiguro1995}, in 1995, implemented an IN-based approach on a six legged robot in order to acquire a gait. Each leg is incorporated with a local immune system (LIN) having four antibodies. Each antibody represents gait behaviors; namely backward, retract, forward and protract. Paratopes and idiotopes of all the antibodies are pre-assigned as to either support or transfer. These LIN are evolved using GA in which "winner takes all" approach is used to select antibodies. Two types of antigen are incorporated; one to input situation, other to represent coordination among local networks. Experimentation is limited to forward movement in which 18 iterations of GA establish a no fall situation. This work by Ishiguro et al. is considered as a first attempt towards physical application of idiotypic network but ad-hoc antibody/antigen allocation restricts such systems to low complexity.

Ishiguro et al. \cite{Ishiguro1996,Ishiguro1997}, in 1996, also proposed a decentralized behavior arbitration scheme to navigate a mobile robot to replenish energy, avoiding obstacles in an arena. It is noted that paratopes are modeled as desirable actions with preassigned definition of action. Idiotopes are modeled as identification numbers that are assigned according to the results of an adjustment mechanism (reinforcement). Antigen are pre-massaged in terms of object information, direction of object and current energy state. Experimental results, however, show limited results of an 18 antibody network that enables the robot to avoid one obstacle to reach the charging station. Antibody selection is done on a "roulette wheel" method. Moreover, the network does not make use of antibody metadynamics but uses an adjustment mechanism to select an idiotope ID.

It should be noted that Jerne's idiotypic theory defines the idiotype in terms of physical connection, like that of a key lock, to identify each other. However, a BIS can open a number of locks with one key. Although this analogy is weak, antigen/antibody allocation in Ishiguro's initial work, however, does not incorporate this phenomenon. Moreover, this approach also avoids use of unstructured environment in simulations. Ishiguro's work was extended by Watanabe et al. \cite{Watanabe1998} to include an off-line innovation function. This innovation function is based on a genetic algorithm with a mixing pot method for crossover operator. It is noted that antibodies are retained as behavior modules. The problem was also extended to add garbage collection behavior in addition to existing obstacle avoidance and energy replenishment behaviors. The drawback of this approach, as well as of Ishiguro's, is the definition of antibodies as behavioral modules. This approach forces one to define behaviors ahead of time with no possibility of behavior evolution.

Michelan and Von Zuben \cite{Michelan2002} improved Ishiguro's model by incorporating a GA-based antibody evolution mechanism. Idiotope is modeled as a set of stimulated antibodies for the network. Antibody affinity is computed on the basis hamming distance evaluation. GA-based adjustment mechanism uses a 40\% crossover and 1\% mutation with an elitist selection. Fitness function is based on number of collected and transferred garbage, recharges and collisions. It should be noted, however, that a BIS has an inherent mechanism to clone antibodies using somatic hypermutation. The above mentioned models use other algorithms for similar purposes. This raises a question on the degree of AIS implementation.

Vargas et al. \cite{Vargas2003} attempted the same garbage collection application using learning classifier system in addition to existing platform provided by Michelan and Von Zuben. This model, named as CLARINET, adds a learning classifier system to classify antigens and antibodies. Antibody structure is, for that purpose, restructured in terms of antecedent and consequent parts to represent paratopes and antibody connections to represent idiotopes. This addition makes antibody network more flexible but requires more computation effort as classifiers are updated both before and after the immune network dynamics. They also implemented the GA-based immune network on Khepra robots \cite{Vargas2003a}.

Krautmacher and Dilger \cite{Krautmacher2004} tried to implement a simplified rescue scenario involving single robot. Antigen are binary coded information of object type and position. Algorithm then uses coordinate transformations for network dynamics in which no metadynamics is incorporated. Rest is same as in Watanabe's approach.

Wang et al. \cite{Wang2007} used the immune network approach of Ishiguro in conjunction with obstacle restriction method (ORM) and reinforcement learning (RL). This application is a single-robot path-planning exercise in which two types of antibodies are defined: one to represent obstacle avoidance behavior and other to seek goal. Antigen is defined as a binary coded data of obstacles/goal in terms of task proximity (near or far). An expression similar to T-cell metaphor, as in Luh et al. \cite{Luh2006}, is also used to help suppress either of the behaviors. This also replaces the need to define $\it{stimulus_2}$. Moreover, cloning or metadynamics is not defined in network structure.

Tsankova et al. \cite{Tsankova2007}, in 2007, applied Ishiguro's network to implement stigmergy based foraging behavior. This work uses different scenarios to collect pucks with single/two robot(s), with one network for goal following behavior and one to pick and drop the pucks. Report also compares the results with Braitenberg's 3C and Q-learning robots. This research does not add to Ishiguro's interpretation but the experimentation establishes a comparative analysis with two well known approaches.

\subsubsection{Whitbrook's Stream:}

Whitbrook et al. \cite{Whitbrook2007} solved the maze-world problem with extensive experimentation using three approaches: RL, RL with simple idiotypic system and RL with full idiotypic AIS. System uses 8 predefined antigen having priority levels assigned to pre-massaged data that translates sensor info into a situation. Sixteen (16) antibodies have predefined behaviors with speed specifications. Idiotopes are fixed while paratopes are predefined that have some adjustment flexibility through reinforcement. Affinity computation is done as in Vargas et al. \cite{Vargas2003}. It is noted that antibody metadynamics is not implemented. Moreover, system uses a-priori information in antigen, antibody and idiotope matrices with limited adjustability of paratopes. Although results show that robot with full feedback performs better in terms of escaping traps by establishing idiotypic network, the system should be able to adjust its internal values automatically either through T-helper cells or through antibody evolution.

Whitbrook further extended her work by incorporating GA to evolve behaviors \cite{Whitbrook2008}. This GA supported long term learning (LTL) combined with short term learning (STL) of idiotypic immune network was tested against STL only approach in \cite{Whitbrook2008a,Whitbrook2010b}. The underlying notion that AIS can only exhibit short term learning can be questioned as it is dependant on system's metadynamics that can be adjusted to retain memory for a longer period.

\subsubsection{Lee-Sim Stream:}

Lee et al. \cite{Lee1997} executed the swarm intelligence task involving multiple robots. Task density, either high, medium, low or nil, is represented as antigen. Antibodies are defined as four behaviors of aggregation, random search, dispersion and homing. Sensors detect the task concentration that is then used in a fuzzy function to output a stimulus-value for Farmer's equation. Resulting concentration is then used to stimulate other robots to do the same task. Metaphors of plasma and deactivated cells are used to incorporate some level of metadynamics. This approach also suffers from inherent problem of a-priori behavior specification. Jun et al. \cite{Jun1999} extended this work by incorporating T-cell metaphor to represent control parameters. This adds another layer in the network and resets the antibody concentrations once an antigen is removed.

\subsubsection{Li-Wang Stream:}

Li and Wang \cite{Li2003} implemented a sheep-and-dog problem using predefined coefficients to compute affinities. Antibody network dynamics is replaced by an algebraic expression that takes into account the usual stimulations and suppressions along with a T-cell function and a linear death rate. Environment is translated into antigen by tabulating positions of dog and sheep in a matrix $X$. Antibody matrix $Y$ also has previous information of actions corresponding to each entry in antigen matrix $X$. Only five actions are possible. This research does not make any distinction between sheep and dog in terms of their embodiment. Moreover, the network uses a manual mechanisms to perform coefficient selection that limits its adaptability.

Duan et al. \cite{Duan2005} extended the work of Li and Wang to perform a predator-prey experiment with 2 predators and one prey, each having a small antibody network that can communicate with each other except in pursuit domain. Antigen is of two types: one has environment information in terms of position data and other has communication signal. Two different antibody structures are implemented for predator and prey robots. Predator has six actions to arbitrate from while prey has three behaviors to choose from, on the basis of synthesized immune network as in Li and Wang \cite{Li2003}.

\subsubsection{Luh's Work:}
Luh has presented three different applications that use different immune metaphors. His approach is based on real data representation scheme. In 2002, he with Cheng \cite{Luh2002} presented a food foraging application that uses APC module to assess the environment and T-cells as a RL mechanism. T-helper-cells are used here as an adaptive critic. Luh et al. \cite{Luh2006}, in 2006, implemented a robot soccer application using immune network. Antigen is sensor information that is mapped to have three components, one is distance between ball and goal, second is distance between ball and robot and third is crowd data. Each of these components corresponds to a fuzzy function to find affinity value. Average of all three affinities, through fuzzy membership functions, is computed in terms of a 6x6 affinity matrix. A T-cell function is incorporated that acts as a reinforcement. Antibody metadynamics is not implemented since there is no repertoire maintained as memory. Zhang and Lu \cite{Zhang2008} reproduced this approach using four antibodies instead of six for each robot.\\

Luh and Liu \cite{Luh2008}, in 2008, solved the robot navigation problem using reactive-IN approach with fused data representation. Antigen is a vector of azimuthal angle of goal, distance information of each sensor and sensor location on the robot periphery. Antibodies are defined as steering directions $(\theta_i)$. Stimulation and suppression due to antibody-antibody interactions is defined as cosine of difference between respective antibodies. Stimulation due to antigen interaction is defined in terms of attractive/repulsive forces of goal seeking and obstacle avoidance. In order to escape robot from trapping in local minima, an adaptive virtual target method is also used. The weighing mechanism of attractive/repulsive forces is manual. Therefore, it is not clear that how a robot manages to arbitrate the two behaviors.

Dehuai et al. \cite{Dehuai2008, Xuemei2009} modified the work of Wang et al. \cite{Wang2007} by defining antigen in terms of task density (high, low or none) and combining antibody structure in one representation. It is also noted that Farmer's equation is not solved by an ODE solver but antibody concentration rate is related directly to behavior modules. Moreover, hamming distance is replaced with Luh's expression of $cos(\Delta {\theta})$ to define antibody stimulation and suppression. This application is also a single robot path planning exercise in which two types of antibodies are defined: one to represent obstacle avoidance behavior and other to seek goal.

\subsubsection{Non-Farmer Approach}

Mitsumoto et al. \cite{Mitsumoto1996, Mitsumoto1997} presented an IN-based approach to control population of multiple robots according to assigned task of load transfer from one station to two storage docks. Each task assignment is treated as an antigen that disturbs the existing population of robots. Algorithm then reconfigures to attain new stability (homeostatic state) by sharing message-antigen with other robots. Each robot, treated as a B-cell, has predefined modules to set global states and behavior strategy. Resultantly, the network is limited only to regulate robot population.

Sathyanath and Sahin \cite{Sathyanath2002} and Opp and Sahin \cite{Opp2004} used a mine detection task to perform single objective task with fixed number of robots using a non-idiotypic approach. Antigen are modeled as mine locations whereas antibodies are defined as robots. The communication between antibodies is a network that provides antigen-locations. Robots are, resultantly, stimulated to move toward the mine in order to defuse it. Suppression is implemented when no antigen is detected and results in random movement. This is unlike Farmer's interpretation of Jerne's idiotypic network theory that ensures communication even in absence of antigen.

Generally, it is because of this notion that cells within an immune network can recognize each other, in addition to recognizing antigens, IN-approach is applied on mobile robotic systems. Any change in environment is detected as an antigen. Possible steering directions/behavior-modules are represented as antibodies. The most stimulated antibody, resulting from immune network approach and supplementary methodologies, is fed to the system as an actuation signal.


\subsection{Robotic Applications using Danger Theory:}

Danger theory is a newer definition of BIS working and, therefore, very few robotic applications are reported in this category. Dendritic cell algorithm by Greensmith \cite{Greensmith2006,Greensmith2008,Greensmith2009} incorporates only the working of antigen presenting cells within a BIS and, therefore, requires other immuno functions to fully implement a three signal model.

Oates et al. \cite{Oates2007} used DCA in a mobile robotic security application for classification purpose. Augmenting the subsumption architecture, the robotic DCA is implemented as a stand-alone behavioral module. DCA processes the sensor data as antigen and generates signals that are either safe, dangerous or PAMP \cite{Aickelin2003}. The output of DCA provides a base for subsuming the behavioral modules. It is to be noted that in subsumption architecture, there is a disagreement among various behavioral modules e.g. react to bumpers, recover from stall, avoid obstacles and explore. Brooks \cite{Brooks1986} suggests that this can be solved by allowing components at one level to subsume components at a lower level. It is because of this reason the approach is called subsumption architecture. This application, however, is a classification problem that does not fully incorporate the behaviors necessary for navigation through a maze.

Prieto et al. \cite{Prieto2008} implemented a preliminary work that uses DT on a metaphorical level only and lacks the necessary mathematical interpretations. This application is a soccer goalkeeper strategy in which APCs are ID of predefined strategy whereas antigen is detection of opponent and ball in the home side. Signal one, two and three correspond to respective closeness of ball to the goal.

\subsection{Robotic Applications using Other Approaches:}
This section lists the robotic applications that do not fall in a single BIS category, lack the basics of a particular definition, and/or use other AIS definitions. Xiong et al. \cite{Xiong2007} implemented a multi-robot system that maps an environment on the basis of market approach. The function of immunity is limited only to optimize the strategy to select goal points during exploration. Auxiliary functions are used for other operations like Bayes theorem for data fusion and diffusivity for robot distribution in arena.

The approach of Yuan et al. \cite{Yuan2009} claims to combine artificial potential field (APF) method of Khatib \cite{Khatib1986} with Jerne's idiotypic network theory but fails to specify antibody dynamics. Representation scheme is binary whereas obstacles and goal are separately coded into it. It also uses antibody vitality along with a learning strategy to execute path planning task.

\section{Findings:}

\subsection{On using AIS}

It is important to understand that some aspects of BIS are still being investigated. Clonal selection theory, the oldest of considered theories, focused on one signal approach in which antigen binds with receptors on a B-cell. Danger theory, the newest in self-nonself approach of BIS, follows a three signal approach as shown in fig. \ref{figure_DT}. Therefore, year of publication is important to evaluate an application in terms of validity of its corresponding BIS explanation at that time. There exists a researcher's dilemma that what to take and what to leave in order to solve a problem, especially when a set of theories explain different aspects of a complex phenomenon. It is of no use that the whole BIS is replicated to solve a relatively simple problem but, at the same time, one should not fall victim to a single aspect of BIS as well.

Literature indicates that some auxiliary functions or subsystems are also required to support the core algorithm, like reinforcement learning, fuzzy systems and/or genetic algorithms. It would be more appropriate to use a computational equivalent from BIS, if available. For example, nature uses somatic hypermutation to evolve antibodies but some researchers have used GA instead (e.g. \cite{Michelan2002,Vargas2003}). It would have been logical to use what nature has chosen for a particular purpose. Similarly, some researchers have used RL (e.g. \cite{Ishiguro1997,Wang2007}) when nature uses similar approach of T-helper-cells. It is, therefore, identified that the degree of biological inspiration can be deeper than some applications show it to be. Moreover, there should be investigations to establish that BIS-inspired auxiliary function(s) can be as effective as some other.

Following subsections indicate the findings pertaining to afore-mentioned AIS approaches as well as the robotic applications.

\subsubsection{Using Clonal Selection}
Wang \& Hrisbrunner \cite{Wang2003} and Li et al. \cite{Li2005} used GA-based crossover operators in their applications. In nature, however, BIS does not use crossover. Without sounding like a purist, this raises a question on validity of such auxiliary functions. Especially when it has been established that CS-algorithm and it variants are equally effective, if not better, in optimization tasks (e.g. \cite{White2003}). A framework for establishing convergence of immune algorithm is also presented in \cite{Cutello2007} and a comparative analysis is presented in \cite{Cutello2010} for various test functions. The representation of antigen and antibody is not explicitly defined and justified in some of applications. The benefit of using a robotic application lies in its embodiment and it should be reflective in corresponding representation schema.

\subsubsection{Using Idiotypic Network}
\begin{figure}[ht]
\begin{center}
\includegraphics[width=3.2in, height=2.5in]{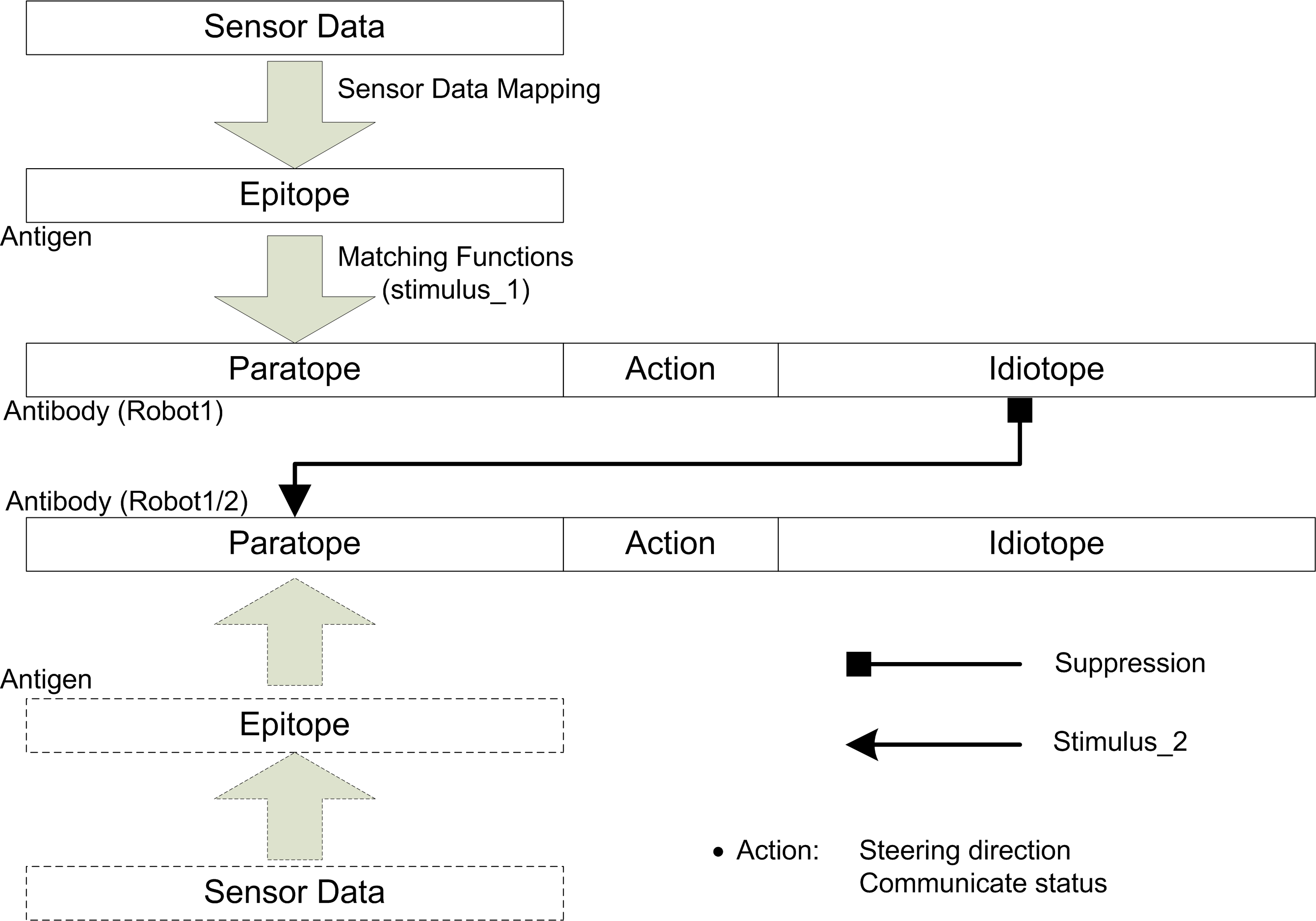}
\end{center}
\caption{A generalized structure of Antigen-Antibody and Antibody-Antibody interactions in an idiotypic network }
\label{figure_Ag_Ab}
\end{figure}

The major chunk of publications fall under the idiotypic network theory. A generic structure of IN is shown in fig. \ref{figure_Ag_Ab} in context of idiotypic connections in a typical robotic application. Antibodies are either evolved or generated by evaluating affinity functions. Mapping schema, affinity definitions and antibody specifications vary from application to application. The question, however, is to justify use of a particular theory (idiotypic theory in this case) and to what extent a theory is applied.

Major criticism on Jerne's idiotypic network theory is in relation to the size of its network \cite{Langman1986}. This criticism arises from a basic argument that how can every antibody recognize every other antibody in a possible network of millions of cells (with current estimates of more than $10^{12}$ lymphocytes). These arguments are not countered in the observed robotic applications. It can, however, be argued that such applications do not require a network of millions of antibodies. In case of behavior arbitration, the network only requires a handful of antibodies. In case of multi-robot applications where robots are generally modeled as B-cells, the size of network does not exceed because of inherent limitations of cost and size of arena.

Farmer's expression of antibody concentration rate (eq. \ref{eq_Farmer1}) is common to every application within idiotypic network theory, barring a few exceptions. It provides an explicit notation to stimulations and suppressions among antibodies and antigens. It, however, has different implementations in different research streams. Apart from conventional approach to use an ODE solver, a discretized version using bilinear transform is employed in Krautmacher \& Dilger \cite{Krautmacher2004} and an algebraic equivalent, named as synthesized immune network, is used in Li-Wang stream. The choice of a particular implementation strategy or a solver is normally dictated by computational requirements of an application. Recently, theoretical issues relating to different AIS algorithms are being raised. The small size of immune network in case of robotic applications induces an effect of discreteness, resulting in difficulties to analyze using standard techniques \cite{Timmis2008}.

Ishiguro-Watanabe stream primarily shares same network structure but differs in the usage of auxiliary/support functions. It  uses reinforcement learning, genetic algorithms, learning classifiers, etc. as auxiliary functions and does not establish a clonal selection in support to the network. Jerne's theory, on the other hand, is built on clonal selection theory. A network-alone approach, like that of Ishiguro-Watanabe stream, reduces such system to a reactive one that has no evolution resulting from cloning \& hypermutation. It perhaps is a result of behavior-arbitration/action-selection approach in which there are only a few preprogrammed behaviors to choose from. Li-Wang and Luh's streams also lack such implementations.

It is important to devise a way to represent environment into antigen $(A_g)$. Similarly, paratopes and idiotopes of antibodies $(A_i)$ should be encoded in such a manner that stimulation-suppression network can be incorporated. Representation of antigen and antibody can be binary (e.g. in Ishiguro-Watanabe stream) or real (e.g. other streams in IN-approach). A binary representation scheme such as that of Ishiguro-Watanabe stream structures the external \& internal data along with the binary coded epitope. Similarly, paratopes \& idiotopes on antibodies are binary strings alongside the pre-programmed actions. Subsequent affinity computations are based on Hamming distances as shown in tables \ref{table_IN_review1} \& \ref{table_IN_review2}. Real data representation is employed in rest of streams and resulting affinities are computed on the basis of euclidean distance, strength-of-match approach or string matching. Ishiguro-Watanabe mode of representation has a problem that its network does not have any information of the robot body but at the same time has a more generic structure. Luh's approach, however, makes use of fused data representation that includes some information of location of sensors within the network. Selection of representation schema in AIS-based robotic applications is, therefore, tricky in terms of inclusion of robot embodiment in network and a consequent loss of generality.

Antibody death is also an important factor in Farmer's expression. A constant antibody death rate, as employed by all the applications in this category, does not serve a purpose when no memory is maintained or when antibody metadynamics is skipped in implementation.

\subsubsection{Using Newer Definitions}
Danger theory can be considered as an extension of self-nonself models. DCA is based on one aspect of this theory that relates APCs to their maturity on the basis of danger/stressed signals in the system. Resultantly, DCA limits itself to the initiation of immuno responses because the theory itself puts limits on that. It can be used in behavior arbitration on the basis of environment contextualization but then it should be supplemented with B-Cell and T-Cell algorithms to complete a three signal immuno function \cite{Aickelin2003}. Only one application is reported in literature that uses DCA in a robotic application. There is a lot of potential in terms of using innate immunity in conjunction with its adaptive counterpart. As a starting suggestion, conflicting objectives during robot navigation can be tested with DCA which is currently limited to static data. Moreover, fuzzy weighing in some instances can be replaced with a DCA to co-stimulate different behavioral modules. The role of T-cells in helping B-cells can also be further refined to a level of developing adaptive critic as well.

\subsection{On immuno-inspired robotics}
Current trends in robotics have migrated from reactive paradigm to hybrid and probabilistic robotics in order to counter uncertainties in sensing and modeling \cite{Thrun2005}. Moreover, single-robot applications have gathered more robots to implement swarm intelligence \cite{Aazahin2005}. Heterogeneous mobile robotic systems, a new trend, involves using robots of different capabilities performing jobs in unstructured environment. Most of the reported immuno-inspired applications, however, involve either single robot or multiple robots of same type. Even in case of Li-Wang stream where predator \& prey experiment is implemented, no distinction is made between predator and prey. In classical predator-prey models, a predator is embedded with higher sensitivities and a prey is modeled with higher actuation capabilities. For example, a predator may be given better vision and a prey may have higher speeds to escape an attack. This approach makes such systems to be heterogeneous mobile robotic systems.  This heterogeneity requires a generalized representation scheme that can handle robots of varying capabilities in terms of their sensing and actuation.

Reported applications also limit themselves in terms of using predefined behaviors to arbitrate from. This poses a problem in case of employing heterogeneous mobile robots because each robot would then require programming of different behavioral modules ahead of time. Ideally, intelligence should emerge irrespective of hardware configuration of robots. This leads us to opt for behavior evolution rather than conventional behavior arbitration because coupling antibodies to predefined actions or behavior-modules stops inclusion of new behaviors.

The ad-hoc manner in which these applications and their outcomes are reported makes us suggest benchmarking. Although robotics is ever-changing, there are a number of scenarios and datasets available. Nowak et al. \cite{BRICS2010} presented a detailed account of related benchmarks as part of the European research project BRICS. The benchmarks should be implemented and compared with other established techniques. Alternatively, some metrics related to utility, cost and reliability should be implemented to support an algorithm.

Robot trapping in a local minima is the most common drawback of using reactive approach. In this context, many trap-escaping schemes have been tried and investigated. Potential field method \cite{Hui2008,Mabrouk2008}, numerical potential field method \cite{Barraquand1992}, virtual target method \cite{Zou2003}, virtual force field method \cite{Borenstein1989}, vector field method \cite{Borenstein1991} are some of the methods that are used to help local minima recovery in robot navigation scenarios. Immuno-inspired methods are also being designed to handle the issue because of adaptive nature of these algorithms e.g. virtual target method by Luh and Liu \cite{Luh2008} but simple shaped-arenas (e.g. U, W \& X) do not fully establish the trap-escaping capabilities of underlying immunity-based technique.

Most of reported literature uses simulations and do not implement the algorithms on real systems. It should be noted that there are issues pertaining to non-holonomic nature of most robotic platforms. In simulations, it is much easier to implement a robot as a dot, irrespective of its dynamics. Similarly, the detection of obstacles, walls or targets is difficult and pose a lot more implementation issues.


\section{Conclusions:}
We have reviewed the literature on immuno-inspired robotic applications. In advent of newer definitions of BIS and current trends in robotics, it is important to categorize these applications in terms of underlying immune definitions, computational details and deficiencies and consequently point towards future directions. It is concluded that a deeper biological inspiration is required because a single aspect of AIS may not be sufficient to incorporate a successful robotic system. Auxiliary functions should be taken from their computational equivalents within BIS, where available. It is important to highlight here that an AIS can be \emph{all-encompassing}; one that combines innate and adaptive immune systems by employing functions of phagocytes, dendritic cells, T \& B lymphocytes, etc. An AIS has functions that provide a distributed network structure like idiotypic network, reinforcement learning like T-cell algorithms, evolutionary mechanism like somatic hypermutation, short term learning like metadynamics and weighted sum of attractive/repulsive forces like dendritic cell algorithm. A two layered approach can be one of the solutions where one layer corresponds to antigenic data and the other to environment contextualization in terms of safe or dangerous signals.

A network-alone approach reduces robotic system to a reactive one that has no evolution resulting from cloning \& hypermutation. Current trends, on the other hand, are more inclined towards behavior evolution rather than behavior arbitration. It is also concluded that with a deeper BIS inspiration it is possible to add stochastic nature of clonal selection to the deterministic approach of idiotypic network. The benefit of using robots as an application is in its embodiment. Fear of unknown environment can be reduced by knowing something about robot. The information of sensors locations and system dynamics can, therefore, be a part of representation schema.

Selection of a particular robotic application is also important. \emph{Search and rescue} scenario involving heterogeneous robots offers a comprehensive application that uses different robot configurations to accomplish a wide variety of tasks, ranging from single robot navigation through obstacles to multi-robot coordinated navigation in rescue. Robot taxonomy is important aspect to be considered as well, especially when one wishes to develop a general algorithm for a number of robot platforms. It has also been identified that benchmark problems should also be tested to validate an algorithm. We also emphasize the need of real experiments to minimize the reality gaps between simulations and actual systems (e.g. issues pertaining to system's uncertainties, holonomicity, local minima recovery, conflict resolution etc.).

This review also offers an insight into older work and presents a critique to individual as well as group works in previous sections. It is evident that AIS-based robotic applications have helped in establishing immunological computations as a successful approach but there exist a number of voids that are needed to be filled up, both in terms of theory and experimentation. There are a number of new directions that can be investigated. A combined framework of neural networks, endocrinal systems and immune systems can be implemented to establish a homeostasis in an embodied system. Another novel idea can be to vaccinate a simple robot with specific antibodies from a specialist/healthy robot. This approach can be used in situations where an untrained robot is desired to learn faster through vaccination. A robot immune system can also be implemented that regulates/repairs its internal organs and vital functions, unlike the conventional approach of using an AIS as a navigation scheme only.


\pagestyle{empty}
\begin{landscape}
\scriptsize
\topcaption{Review of Clonal Selection Based Robotic Applications}
\label{table_CS_review1}

\tablefirsthead{\toprule
\multirow{2}{*}{}
& Antigen & Antibody & Affinity & Auxiliary Functions & Experiment Specifications \\[0.5ex]\midrule \midrule }

\tablehead{
& & & & & Table continued \\ [0.5ex] \toprule
& Antigen & Antibody & Affinity & Auxiliary Functions & Experiment Specifications \\[0.5ex]\midrule \midrule }

\tablelasthead{ & & & & & Table continued \\ [0.5ex] \toprule
& Antigen & Antibody & Affinity & Auxiliary Functions & Experiment Specifications \\[0.5ex]\midrule \midrule }

\tabletail{\midrule & & & & & continued on next page \\ [0.5ex] \midrule}
\tablelasttail{& & & & & Table ended \\ [0.5ex] \midrule \bottomrule}

\renewcommand\arraystretch{1.6}
\begin{xtabular}{p{2cm} @{}c c c c p{4cm}}

%
\multirow{3}{*}{Hu}
& --- & Grid points of Robot path, & $\displaystyle{fitness = \frac{A}{f}}$, $\displaystyle{f =\sum_{i=1}^{N}\left(d_i + c \beta_i\right)}$ & Roulette wheel selection,& Robot path planning, \\
& & $\displaystyle{X = \left[a_o,a_1, \cdots ,a_n\right]}$ & $\displaystyle{d_i = \sqrt{(x_i-x_{i-1})^2+(y_i-y_{i-1})^2}}$ & Mutation, insertion \& &\\
& & $a_o$ = start pt, $a_n$ = end pt & $\displaystyle{\beta_i = \sum_{j=1}^{M} \alpha_j}$ (if obstacle, $0$ otherwise) & deletion operators&\\
\midrule

\multirow{3}{*}{Wang \&}
& Problem & Possible solution & $\displaystyle{F_i = \nu_i\left[1+\zeta \frac{\nu_i}{max(\nu)}+\delta D_s \left(1-\frac{\nu_i}{max(\nu)}\right)\right]}$ & GA: crossover \& mutation & IMEA for robot path planning, \\
Hirsbrunner& & & $\displaystyle{D_s = \frac{\sum_{i}^{}(\theta.max(\nu) \le \nu_i \le max(\nu))}{\mu+\phi}}$ & &no hypermutation or cloning\\
& & & $\displaystyle{\nu_i = \frac{M.N}{\sum_{i}^{N} \sqrt{(x_{pi}-x_{qi})^2+(y_{pi}-y_{qi})^2}}}$& &\\
\midrule

\multirow{3}{*}{Li et al.}
& --- & Modeled as chromosomes & $\displaystyle{f = \sum min(1-O_{ij},1-E_{ij}+w_1\sum\delta_{ij}+w_2\sum\zeta_{ij})}$& GA: crossover & Single robot performing\\
& & $\displaystyle{X = \left[X_1,X_2,\cdots,X_N \right]}$ & $O_{ij} \rightarrow$ occupancy& Vaccination operator &concurrent mapping \&  \\
& & where $\displaystyle{X_j = \left[\Delta d_j, \Delta \theta_j \right]}$& $E_{ij} \rightarrow$ empty & & localization\\
\midrule

\multirow{3}{*}{Hur}
& Binary coded & Binary coded & $\displaystyle{f = \frac{1}{\left[1+\sum_{i}^{k}w_i.n_i\right]}}$ & Memory updation, & Multi robot bomb disposal \\
&$\displaystyle{\{0,1\}^2}$ & $\displaystyle{\{0,1\}^3}$, Action & $\displaystyle{k \rightarrow}$ no. of soft constraints & Cloning with & system with three robots \\
& & & $\displaystyle{n_i \rightarrow}$ no. of constraints within an $A_b$ & $\displaystyle{N_c = \sum_{i=1}^{n} round (\frac{\beta . N}{i})}$ & (Scanner, inspector, diffuser)\\
\midrule


\end{xtabular}
\normalsize
\end{landscape}


\pagestyle{empty}
\begin{landscape}
\scriptsize
\topcaption{Review of Immune Network Based Robotic Applications in Terms of Definitions}
\label{table_IN_review1}
\tablefirsthead{\toprule
\multirow{2}{*}{}
& Antigen & Antibody & Affinity & $A_b$ Selection & Experiment Specifications \\ [0.5ex]
& $(A_g)$ & $(A_b)$ & & &\\ [0.5ex] \midrule \midrule }
\tablehead{
& & & & & Table continued \\ [0.5ex] \toprule
\multirow{2}{*}{}
& Antigen & Antibody & Affinity & $A_b$ Selection & Experiment Specifications \\ [0.5ex]
& $(A_g)$ & $(A_b)$ & & &\\ [0.5ex] \midrule \midrule }
\tablelasthead{ & & & & & Table continued \\ [0.5ex] \toprule
\multirow{2}{*}{}
& Antigen & Antibody & Affinity & $A_b$ Selection & Experiment Specifications \\ [0.5ex]
& $(A_g)$ & $(A_b)$ & & &\\ [0.5ex] \midrule }
\tabletail{\midrule & & & & & continued on next page \\ [0.5ex] \midrule}
\tablelasttail{& & & & & Table ended \\ [0.5ex] \midrule \bottomrule}

\renewcommand\arraystretch{1.55}
\begin{xtabular}{p{2cm} @{}c c c c p{4cm}}

\multirow{2}{*}{Ishiguro et al.}
& Pre-massaged Data & Behavior modules (fixed), & Adjustment Mechanism & Roulette Wheel & Single robot, \\
&\includegraphics[scale=0.55]{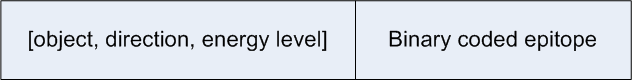}&\includegraphics[scale=0.55]{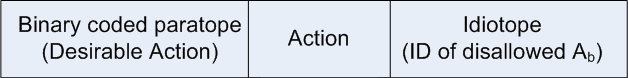}& & &Seek energy and avoid obstacles\\
\midrule

\multirow{2}{*}{Watanabe et al.}
& Binary coded, pre-massaged & Behavior modules (fixed), & Hamming distance & Roulette Wheel & Single robot,\\
&\includegraphics[scale=0.55]{Ag1.png}&\includegraphics[scale=0.55]{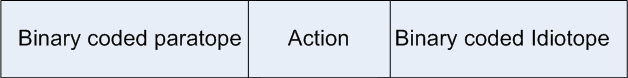}& & & Collect garbage, seek energy and avoid obstacles\\
\midrule

\multirow{2}{*}{Michelan and}
& Binary coded, pre-massaged & Behavior modules (fixed), & Hamming distance & Roulette Wheel & Single robot,\\
Von Zuben &\includegraphics[scale=0.55]{Ag1.png}&\includegraphics[scale=0.55]{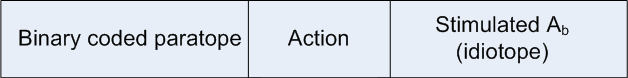}& & & Collect garbage, seek energy and avoid obstacles\\
\midrule

\multirow{2}{*}{Vargas et al.}
& Ternary, through classifier & Behavior modules (2 part paratope), & Hamming distance & Roulette Wheel & Single robot w/ learning classifier\\
&\includegraphics[scale=0.55]{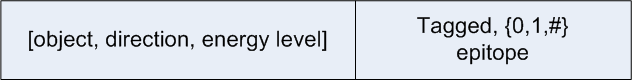}&\includegraphics[scale=0.55]{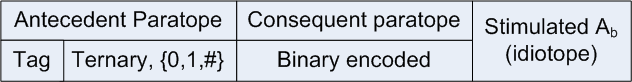}& & & Collect garbage, seek energy and avoid obstacles\\
\midrule

\multirow{2}{*}{Krautmacher}
& Binary encoded, variable & Behavior modules, & Hamming distance & --- & Single robot,\\
and Dilger &\includegraphics[scale=0.55]{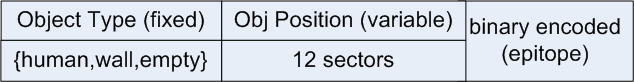}&\includegraphics[scale=0.55]{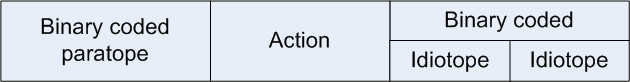}& & & Simple rescue scenario\\
\midrule

\multirow{2}{*}{Wang et al.}
& Binary encoded, pre-massaged, & Binary coded (separate) & Hamming distance & --- & Single robot path planning,\\
&\includegraphics[scale=0.55]{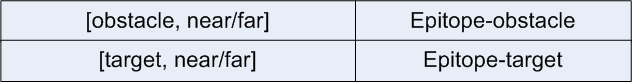} & \includegraphics[scale=0.55]{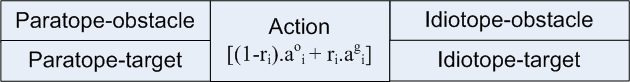}& & & Seek goal and avoid obstacles\\
\midrule

\multirow{2}{*}{Tsankova et al.}
& Binary coded, pre-massaged & Behavior modules (fixed), & Hamming distance & Roulette Wheel & Single/two robot(s),\\
&\includegraphics[scale=0.55]{Ag1.png}&\includegraphics[scale=0.55]{Ab2.png}& & & Stigmergy based foraging behavior with two networks\\
\midrule

\multirow{2}{*}{Whitbrook et al.}
& 8 predefined situations & 16 predefined actions, & Strength of match & feedback & Robot navigation in maze world,\\
&\includegraphics[scale=0.55]{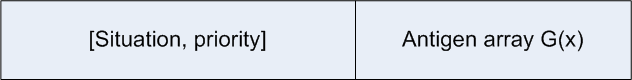}&\includegraphics[scale=0.55]{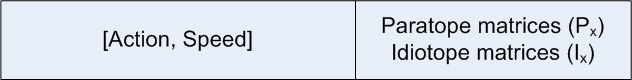}& by reinforcement scores & & fixed idiotpe and adjustable paratope matrices\\
\midrule

\multirow{2}{*}{Lee and Sim}
& Real, classification of task density & Real, action strategy, & Combined stimulation & winner take all & Multi robot system,\\
&\includegraphics[scale=0.55]{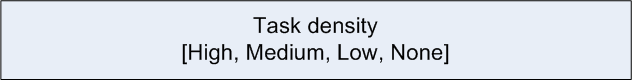}&\includegraphics[scale=0.55]{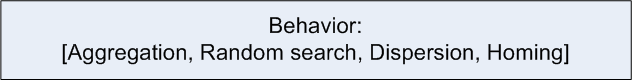}& and suppression & & Mine detection using same robots\\
\midrule

\multirow{2}{*}{Jun et al.}
& Real, classification of task density & Real, action strategy, & Combined stimulation & T-cell assisted & Multi robot system,\\
&\includegraphics[scale=0.55]{Ag6.png}&\includegraphics[scale=0.55]{Ab8.png}& and suppression & & Mine detection using same robots\\
\midrule

\multirow{2}{*}{Li and Wang}
& Real, Environment coding & Real, Action strategy, & String matching & Synthesized & Dog and Sheep problem,\\
&\includegraphics[scale=0.55]{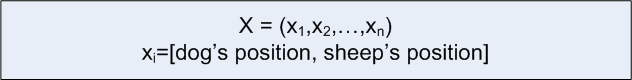}&\includegraphics[scale=0.55]{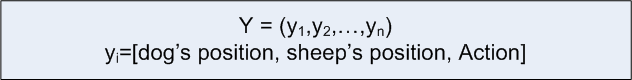}& & immune network & same type of robots\\
\midrule

\multirow{2}{*}{Duan et al.}
& Real, Environment coding & Separate action strategies, & String matching & Synthesized & Predator and prey problem,\\
&\includegraphics[scale=0.55]{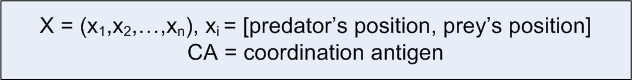}&\includegraphics[scale=0.55]{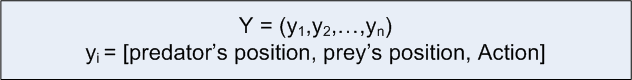}& & immune network & same type of robots\\
\midrule

\multirow{2}{*}{Luh et al.}
& Real data & 6 predefined behaviors, & Average from FIS & winner take all & Robot soccer application,\\
&\includegraphics[scale=0.55]{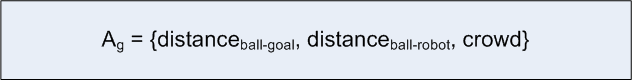}&\includegraphics[scale=0.55]{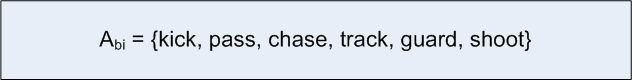}&in a 6x6 matrix & & All 4 robots are of same type\\
\midrule

\multirow{2}{*}{Luh and Liu}
& Real fused data, & Steering directions, & --- & winner take all & Single robot navigation,\\
&\includegraphics[scale=0.55]{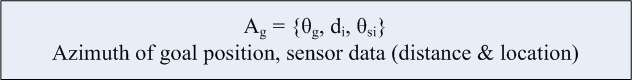}&\includegraphics[scale=0.55]{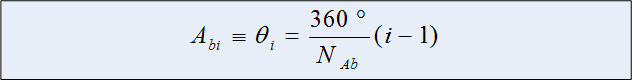}& & & using local reactive approach\\
\midrule

\multirow{2}{*}{Dehuai et al.}
& Real,& Real, Steering directions, & --- & $a_i=(1-\gamma_i)a_i^o + \gamma a_i^g$ & Car like robot,\\
&\includegraphics[scale=0.55]{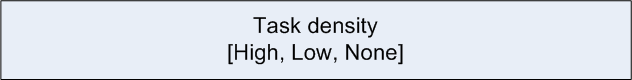}&\includegraphics[scale=0.55]{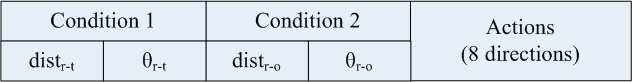}& &$\gamma$ through distance & path planning with obstacle avoidance\\
\midrule

\end{xtabular}
\normalsize
\end{landscape}


\pagestyle{empty}
\begin{landscape}
\topcaption{Details of Network Structure in IN-Based Robotic Applications}
\label{table_IN_review2}
{\centering
\scriptsize

\tablefirsthead{
\toprule
\multirow{2}{*}{}
& Suppression & Stimulus 1 & Stimulus 2 & Cloning & Metadynamics & Auxiliary Functions \\ [0.5ex]
& $(A_b-A_b)$ & $(A_b-A_b)$ & $(A_g-A_b)$ & & &\\ [0.5ex] \midrule \midrule }

\tablehead{ & & & & & & Table continued \\ [0.5ex] \toprule
\multirow{2}{*}{}
& Suppression & Stimulus 1 & Stimulus 2 & Cloning & Metadynamics & Auxiliary Functions \\ [0.5ex]
& $(A_b-A_b)$ & $(A_b-A_b)$ & $(A_g-A_b)$ & & & \\ [0.5ex] \midrule \midrule }

\tablelasthead{ & & & & & & Table continued \\ [0.5ex] \toprule
\multirow{2}{*}{}
& Suppression & Stimulus 1 & Stimulus 2 & Cloning & Metadynamics & Auxiliary Functions \\ [0.5ex]
& $(A_b-A_b)$ & $(A_b-A_b)$ & $(A_g-A_b)$ & & &\\ [0.5ex] \midrule \midrule }

\tabletail{\midrule & & & & & & continued on next page \\ [0.5ex] \midrule}
\tablelasttail{& & & & & & Table ended \\ [0.5ex] \midrule \bottomrule}

\renewcommand\arraystretch{2}
\begin{xtabular}{p{2cm} @{}c c c c c p{3.5cm}} 

Ishiguro et al.
& $\displaystyle{\frac{T_p^{Ab_i} + T_r^{Ab_j}}{T_{Ab_j}^{Ab_i}}}$
& $\displaystyle{\frac{T_r^{Ab_j} + T_p^{Ab_i}}{T_{Ab_i}^{Ab_j}}}$
& Hamming distance & --- & --- & Reinforcement learning to compute $T_x$ \\
\midrule

Watanabe et al.
& $\displaystyle{F \left[\sum_{k=1}^{L}(w_k \overline{I_i(k) \oplus P_j(k)})\right]}$
& $\displaystyle{F\left[\sum_{k=1}^{L}(w_k \overline{I_j(k) \oplus P_i(k)})\right]}$
& $\displaystyle{\sum_{j} \left[F \left(\sum_{k=1}^{L}(w_k \overline{E_j(k) \oplus P_i(k)})\right)\right]}$
& --- & GA based & Mixing pot crossover function \\
\midrule

Michelan and Von Zuben
& $\displaystyle{m_{ij} = \frac{T_p^{Ab_i} + T_r^{Ab_j}}{T_{Ab_j}^{Ab_i}}}$
& $\displaystyle{m_{ji} = \frac{T_r^{Ab_j} + T_p^{Ab_i}}{T_{Ab_i}^{Ab_j}}}$
& Hamming distance & --- & --- & GA based adjustment scheme with elitist selection \\
\midrule

Vargas et al.
& $\displaystyle{F \left[\sum_{k=1}^{L}(w_k \overline{I_i(k) \oplus P_j(k)})\right]}$
& $\displaystyle{F\left[\sum_{k=1}^{L}(w_k \overline{I_j(k) \oplus P_i(k)})\right]}$
& $\displaystyle{\sum_{j} \left[F \left(\sum_{k=1}^{L}(w_k \overline{E_j(k) \oplus P_i(k)})\right)\right]}$
& --- & --- & GA based adjustment and learning classifier system \\
\midrule

Krautmacher and Dilger
& $\displaystyle{F \left[\sum_{k=1}^{L}(w_k \overline{I_i(k) \oplus P_j(k)})\right]}$
& $\displaystyle{F\left[\sum_{k=1}^{L}(w_k \overline{I_j(k) \oplus P_i(k)})\right]}$
& Hamming distance $\forall A_g$ & --- & --- & Bilinear transformation for network dynamics \\
\midrule

Wang et al.
& $\displaystyle{\sum_{k=1}^{L}(I_i(k) \oplus \overline{P_j(k)})}$
& $\displaystyle{\sum_{k=1}^{L}(I_j(k) \oplus \overline{P_i(k)})}$
& $\displaystyle{\sum_{k=1}^{L}(E_j(k) \oplus \overline{P_i(k)})}$
& --- & --- & Strengthened learning (RL), Obstacle restriction method \\
\midrule

Tsankova et al.
& $\displaystyle{\sum_{k=1}^{L}(I_i(k) \oplus \overline{P_j(k)})}$
& $\displaystyle{\sum_{k=1}^{L}(I_j(k) \oplus \overline{P_i(k)})}$
& $\displaystyle{\sum_{k=1}^{L}(E_j(k) \oplus \overline{P_i(k)})}$
& --- & --- & \emph{a-priori} antibody idiotopes and paratopes \\
\midrule

Whitbrook et al.
& $\displaystyle{P[x_{w1}, y_m]I[x_i, y_m]H_i}$
& $\displaystyle{(1 - P[x_i, y_p]) I[x_{w1}, y_p]H_i}$
& $\displaystyle{P[x_i,y_j]G(x_i)_j}$
& $A_b$ concentration & GA (ext.) & RL: $\displaystyle{P[x_w,y_d](t+1)}$ = $\displaystyle{max(0,P[x_w,y_d](t)+rf_{t+0.5})}$ \\
\midrule

Lee and Sim
& Pre-computed $\gamma_{ij}$
& Pre-computed $\gamma_{ij}$
& $g_i$ from FIS
& Strategy transfer & --- & FIS with task density as input  \\
\midrule

Jun et al.
& Pre-computed $m_{ij}$
& Pre-computed $m_{ij}$
& $g_i$ from FIS
& Strategy transfer & --- & T-cell concentration: $\eta(1-g_i)A_i$ \\
\midrule

Li and Wang
& $\displaystyle{\sum_{k=1}^{N}m_{ik}g_k}$
& $\displaystyle{\sum_{j=1}^{N}m_{ij}g_j}$
& Affinity: $g_j$
& --- & --- & T-cell: $\displaystyle{\eta(1-g_i)}$ \\
\midrule

Duan et al.
& ---
& $\displaystyle{\sum_{j=1}^{N}m_{ij}g_{ij_k}}$
& $g_j$ \& $CA_{ik}$
& --- & --- & --- \\
\midrule

Luh et al.
& --- & --- & $\displaystyle{\frac {m_{cw}+m_r+m_b}{3}}$
& --- & --- & FIS is used for ${m_{cw}, m_r, m_b}$ \\
\midrule

Luh and Liu
& $\displaystyle{\cos (\theta_j - \theta_i)}$
& $\displaystyle{\cos (\theta_i - \theta_j)}$
& $\displaystyle{f_{target} + f_{obstacle}}$
& --- & --- & Virtual target method for local minima recovery \\
\midrule

Dehuai et al.
& $\displaystyle{\left[\cos (\theta_j^o - \theta_i^o),\cos (\theta_j^g - \theta_i^g)\right]}$
& $\displaystyle{\left[\cos (\theta_i^o - \theta_j^o),\cos (\theta_i^g - \theta_j^g)\right]}$
& potential functions
& --- & --- & --- \\
\midrule

\end{xtabular}
\normalsize
}
\end{landscape}


\section{References}





\bibliographystyle{elsarticle-num}
\bibliography{Ref_PhD_prop_IIR}

\begin{thebibliography}{10}
\expandafter\ifx\csname url\endcsname\relax
  \def\url#1{\texttt{#1}}\fi
\expandafter\ifx\csname urlprefix\endcsname\relax\def\urlprefix{URL }\fi
\expandafter\ifx\csname href\endcsname\relax
  \def\href#1#2{#2} \def\path#1{#1}\fi

\bibitem{Dasgupta2009}
D.~Dasgupta, L.~F. Nino, Immunological Computation: Theory and Applications,
  Auerbach Publications, 2009.

\bibitem{Castro2002}
L.~N.~d. Castro, Artificial Immune Systems: A New Computational Intelligence
  Approach, Springer-Verlag, London, 2002.

\bibitem{Hart2008}
E.~Hart, J.~Timmis, Application areas of {AIS}: The past, the present and the
  future, Appl. Soft Comput. 8~(1) (2008) 191--201.

\bibitem{Ghallab2004}
M.~Ghallab, An overview of planning technology in robotics, KI 2004: Advances
  in Artificial Intelligence (2004) 29--49.

\bibitem{Brooks1986}
R.~Brooks, A robust layered control system for a mobile robot, Robotics and
  Automation, IEEE Journal of 2~(1) (1986) 14 -- 23.

\bibitem{Baldassarre2003}
G.~Baldassarre, S.~Nolfi, D.~Parisi, Evolving mobile robots able to display
  collective behaviors, Artificial Life 9~(3) (2003) 255--267.

\bibitem{Tuci2008}
E.~Tuci, C.~Ampatzis, F.~Vicentini, M.~Dorigo, Evolving homogeneous
  neurocontrollers for a group of heterogeneous robots: Coordinated motion,
  cooperation, and acoustic communication, Artificial Life 14~(2) (2008)
  157--178.

\bibitem{Wang2008}
J.~Wang, M.~Lewis, Assessing cooperation in human control of heterogeneous
  robots, in: Proceedings of the 3rd ACM/IEEE international conference on Human
  robot interaction, HRI '08, ACM, New York, NY, USA, 2008, pp. 9--16.

\bibitem{Pallottino2007}
L.~Pallottino, V.~Scordio, A.~Bicchi, E.~Frazzoli, Decentralized cooperative
  policy for conflict resolution in multivehicle systems, Robotics, IEEE
  Transactions on 23~(6) (2007) 1170 --1183.

\bibitem{Powers2010}
M.~D. Powers, Applying inter-layer conflict resolution to hybrid robot control
  architectures, Ph.D. thesis, Georgia Institutue of Technology (2010).

\bibitem{Thrun2005}
S.~Thrun, W.~Burgard, D.~Fox, Probabilistic Robotics, MIT Press, 2005.

\bibitem{Konig2009}
L.~K\"{o}nig, S.~Mostaghim, H.~Schmeck, Online and onboard evolution of robotic
  behavior using finite state machines, in: Proceedings of The 8th
  International Conference on Autonomous Agents and Multiagent Systems - Volume
  2, AAMAS '09, International Foundation for Autonomous Agents and Multiagent
  Systems, Richland, SC, 2009, pp. 1325--1326.

\bibitem{Burnet1959}
F.~M. Burnet, The Clonal Selection: Theory of Acquired Immunity, Vanderbilt
  University Press Tennessee, 1959.

\bibitem{Bretscher1970}
P.~Bretscher, M.~Cohn, A theory of self-nonself discrimination, Science
  169~(3950) (1970) 1042--1049.

\bibitem{Jerne1974}
N.~K. Jerne, Towards a network theory of the immune system., Annales
  d'immunologie 125C~(1-2) (1974) 373--389.

\bibitem{Matzinger2002}
P.~Matzinger, The danger model: a renewed sense of self., Science 296~(5566)
  (2002) 301--305.

\bibitem{Dasgupta2006}
D.~Dasgupta, Advances in artificial immune systems, Computational Intelligence
  Magazine, IEEE 1~(4) (2006) 40--49.

\bibitem{2007}
Understanding the Immune System: How It Works, National Institute of Allergy
  and Infectious Diseases, 2007.

\bibitem{Janeway2001}
C.~A. Janeway, How the immune system works to protect the host from infection:
  A personal view, Proceedings of the National Academy of Sciences 98~(13)
  (2001) 7461--7468.

\bibitem{White2003}
J.~White, S.~Garrett, Improved pattern recognition with artificial clonal
  selection?, Artificial Immune Systems (2003) 181--193.

\bibitem{Garrett2003}
S.~Garrett, A paratope is not an epitope: Implications for immune network
  models and clonal selection, Artificial Immune Systems (2003) 217--228.

\bibitem{Farmer1986}
J.~D. Farmer, N.~H. Packard, A.~S. Perelson, The immune system, adaptation, and
  machine learning, Phys. D 2~(1-3) (1986) 187--204.

\bibitem{Langman1986}
R.~E. Langman, M.~Cohn, The 'complete' idiotype network is an absurd immune
  system, Immunology Today 7~(4) (1986) 100 -- 101.

\bibitem{DeBoer1989}
R.~De~Boer, P.~Hogeweg, Stability of symmetric idiotypic networks—a critique
  of hoffmann's analysis, Bulletin of Mathematical Biology 51~(2) (1989)
  217--222.

\bibitem{Aickelin2003}
U.~Aickelin, P.~Bentley, S.~Cayzer, J.~Kim, J.~McLeod, Danger theory: The link
  between ais and ids?, Artificial Immune Systems (2003) 147--155.

\bibitem{Greensmith2006}
J.~Greensmith, U.~Aickelin, J.~Twycross, Articulation and clarification of the
  dendritic cell algorithm, in: H.~Bersini, J.~Carneiro (Eds.), Artificial
  Immune Systems, Vol. 4163 of Lecture Notes in Computer Science, Springer
  Berlin / Heidelberg, 2006, pp. 404--417.

\bibitem{Twycross2007}
J.~Twycross, Integrated innate and adaptive artificial immune systems applied
  to process anomaly detection, Ph.D. thesis, University of Nottingham (January
  2007).

\bibitem{Luh2008}
G.~C. Luh, W.~W. Liu, An immunological approach to mobile robot reactive
  navigation, Appl. Soft Comput. 8~(1) (2008) 30--45.

\bibitem{Lau2007}
H.~Y. Lau, V.~W. Wong, I.~S. Lee, Immunity-based autonomous guided vehicles
  control, Applied Soft Computing 7~(1) (2007) 41 -- 57.

\bibitem{Hu2008}
X.~Hu, Clonal selection based mobile robot path planning, in: Automation and
  Logistics, 2008. ICAL 2008. IEEE International Conference on, 2008, pp. 437
  --442.

\bibitem{Wang2003}
L.~Wang, B.~Hirsbrunner, An evolutionary algorithm with population immunity and
  its application on autonomous robot control, in: Evolutionary Computation,
  2003. CEC '03. The 2003 Congress on, Vol.~1, 2003, pp. 397 -- 404 Vol.1.

\bibitem{Li2005}
M.~Li, Z.~Cai, Y.~Shi, P.~Gao, A hybrid immune evolutionary computation based
  on immunity and clonal selection for concurrent mapping and localization,
  Advances in Natural Computation (2005) 1308--1311.

\bibitem{Hur2007}
J.~Hur, Multi-robot system control using artificial immune system, Ph.D.
  thesis, University of Texas at Austin (2007).

\bibitem{Chingtham2009}
T.~Chingtham, S.~Nair, Modeling a multiagent mobile robotics test bed using a
  biologically inspired artificial immune system, Multi-Agent Systems for
  Society (2009) 270--283.

\bibitem{Jun1999}
J.-H. Jun, D.-W. Lee, K.-B. Sim, Realization of cooperative strategies and
  swarm behavior in distributed autonomous robotic systems using artificial
  immune system, in: Systems, Man, and Cybernetics, 1999. IEEE SMC '99
  Conference Proceedings. 1999 IEEE International Conference on, Vol.~6, 1999,
  pp. 614 --619 vol.6.

\bibitem{Ishiguro1995}
A.~Ishiguro, S.~Kuboshiki, S.~Ichikawa, Y.~Uchikawa, Gait coordination of
  hexapod walking robots using mutual-coupled immune networks, in: Proc. IEEE
  International Conference on Evolutionary Computation, Vol.~2, 1995, pp.
  672--677.

\bibitem{Ishiguro1996}
A.~Ishiguro, T.~Kondo, Y.~Watanabe, Y.~Uchikawa, Immunoid: An immunological
  approach to decentralized behavoir arbitration of autonomous mobile robots,
  in: PPSN IV: Proceedings of the 4th International Conference on Parallel
  Problem Solving from Nature, Springer-Verlag, London, UK, 1996, pp. 666--675.

\bibitem{Ishiguro1997}
A.~Ishiguro, Y.~Watanabe, T.~Kondo, Y.~Shirai, Y.~Uchikawa, A robot with a
  decentralized consensus-making mechanism based on the immune system, in:
  Autonomous Decentralized Systems, 1997. Proceedings. ISADS 97., Third
  International Symposium on, 1997, pp. 231 --237.

\bibitem{Watanabe1998}
Y.~Watanabe, A.~Ishiguro, Y.~Shirai, Y.~Uchikawa, Emergent construction of
  behavior arbitration mechanism based on the immune system, in: Evolutionary
  Computation Proceedings, 1998. IEEE World Congress on Computational
  Intelligence., The 1998 IEEE International Conference on, 1998, pp. 481
  --486.

\bibitem{Michelan2002}
R.~Michelan, F.~J. Von~Zuben, Decentralized control system for autonomous
  navigation based on an evolved artificial immune network, in: CEC '02:
  Proceedings of the Evolutionary Computation on 2002. CEC '02. Proceedings of
  the 2002 Congress, IEEE Computer Society, Washington, DC, USA, 2002, pp.
  1021--1026.

\bibitem{Vargas2003}
P.~A. Vargas, L.~N. de~Castro, R.~Michelan, F.~J.~V. Zuben, An immune learning
  classifier network for autonomous navigation., in: J.~Timmis, P.~J. Bentley,
  E.~Hart (Eds.), ICARIS, Vol. 2787 of Lecture Notes in Computer Science,
  Springer, 2003, pp. 69--80.

\bibitem{Vargas2003a}
P.~Vargas, L.~de~Castro, R.~Michelan, F.~Von~Zuben, Implementation of an
  immuno-genetic network on a real khepera ii robot, in: Evolutionary
  Computation, 2003. CEC '03. The 2003 Congress on, Vol.~1, 2003, pp. 420 --
  426 Vol.1.

\bibitem{Krautmacher2004}
M.~Krautmacher, W.~Dilger, Ais based robot navigation in a rescue scenario,
  Artificial Immune Systems (2004) 106--118.

\bibitem{Wang2007}
Y.-N. Wang, T.-S. Lee, T.-F. Tsao, Plan on obstacle-avoiding path for mobile
  robots based on artificial immune algorithm, in: D.~Liu, S.~Fei, Z.-G. Hou,
  H.~Zhang, C.~Sun (Eds.), Advances in Neural Networks – ISNN 2007, Vol. 4491
  of Lecture Notes in Computer Science, Springer Berlin / Heidelberg, 2007, pp.
  694--703.

\bibitem{Luh2006}
G.-C. Luh, C.-Y. Wu, W.-W. Liu, Artificial immune system based cooperative
  strategies for robot soccer competition, in: Strategic Technology, The 1st
  International Forum on, 2006, pp. 76 --79.

\bibitem{Tsankova2007}
D.~Tsankova, V.~Georgieva, F.~Zezulka, Z.~Bradac, Immune network control for
  stigmergy based foraging behaviour of autonomous mobile robots, Int. J.
  Adapt. Control Signal Process. 21~(2-3) (2007) 265--286.

\bibitem{Whitbrook2007}
A.~Whitbrook, U.~Aickelin, J.~Garibaldi, Idiotypic immune networks in
  mobile-robot control, Systems, Man, and Cybernetics, Part B: Cybernetics,
  IEEE Transactions on 37~(6) (2007) 1581 --1598.

\bibitem{Whitbrook2008}
A.~Whitbrook, U.~Aickelin, J.~Garibaldi, Genetic-algorithm seeding of idiotypic
  networks for mobile-robot navigation, Proceedings of the International
  Conference on Informatics in Control, Automation and Robotics (ICINCO 2008),
  in print, pp, Funchal, Portugal, 2008~(arXiv:0803.1626).

\bibitem{Whitbrook2008a}
A.~Whitbrook, U.~Aickelin, J.~Garibaldi, An idiotypic immune network as a
  short-term learning architecture for mobile robots, in: P.~Bentley, D.~Lee,
  S.~Jung (Eds.), Artificial Immune Systems, Vol. 5132 of Lecture Notes in
  Computer Science, Springer Berlin / Heidelberg, 2008, pp. 266--278.

\bibitem{Whitbrook2010b}
A.~M. Whitbrook, U.~Aickelin, J.~M. Garibaldi, Two-timescale learning using
  idiotypic behaviour mediation for a navigating mobile robot, Applied Soft
  Computing 10~(3) (2010) 876 -- 887.

\bibitem{Lee1997}
D.~W. Lee, K.~B. Sim, Artificial immune network-based cooperative control in
  collective autonomous mobile robots, in: Robot and Human Communication, 1997.
  RO-MAN '97. Proceedings., 6th IEEE International Workshop on, 1997, pp. 58
  --63.

\bibitem{Li2003}
J.~H. Li, S.~A. Wang, Model of immune agent and application in path finding of
  autonomous robots, in: Proc. International Conference on Machine Learning and
  Cybernetics, Vol.~3, 2003, pp. 1961--1964.

\bibitem{Duan2005}
Q.~Duan, R.~Wang, H.~Feng, L.~Wang, Applying synthesized immune networks
  hypothesis to mobile robots, in: Autonomous Decentralized Systems, 2005.
  ISADS 2005. Proceedings, 2005, pp. 69 -- 73.

\bibitem{Luh2002}
G.-C. Luh, W.-C. Cheng, Behavior-based intelligent mobile robot using an
  immunized reinforcement adaptive learning mechanism, Advanced Engineering
  Informatics 16~(2) (2002) 85 -- 98.

\bibitem{Zhang2008}
Y.~Zhang, T.~Lu, Research on cooperative strategies of soccer robots based on
  artificial immune system, in: Computational Intelligence and Industrial
  Application, 2008. PACIIA '08. Pacific-Asia Workshop on, Vol.~2, 2008, pp.
  656 --659.

\bibitem{Dehuai2008}
Z.~Dehuai, X.~Gang, X.~Cunxi, Y.~degui, Artificial immune algorithm based robot
  obstacle-avoiding path planning, in: Automation and Logistics, 2008. ICAL
  2008. IEEE International Conference on, 2008, pp. 798 --803.

\bibitem{Xuemei2009}
L.~Xuemei, K.-C. Kwong, L.~Jincheng, J.~Liangzhong, Z.~Dehuai, Mobile robot
  path planning based on artificial immune algorithm, in: Advanced Robotics and
  its Social Impacts (ARSO), 2009 IEEE Workshop on, 2009, pp. 1 --5.

\bibitem{Mitsumoto1996}
N.~Mitsumoto, T.~Fukuda, F.~Arai, H.~Tadashi, T.~Idogaki, Self-organizing
  multiple robotic system (a population control through biologically inspired
  immune network architecture), Robotics and Automation, 1996. Proceedings.,
  1996 IEEE International Conference on 2 (1996) 1614--1619 vol.2.

\bibitem{Mitsumoto1997}
N.~Mitsumoto, T.~Fukuda, F.~Arai, H.~Ishihara, Control of the distributed
  autonomous robotic system based on the biologically inspired immunological
  architecture, in: Robotics and Automation, 1997. Proceedings., 1997 IEEE
  International Conference on, Vol.~4, 1997, pp. 3551 --3556 vol.4.

\bibitem{Sathyanath2002}
S.~Sathyanath, F.~Sahin, Application of artificial immune system based
  intelligent multi agent model to a mine detection problem, in: Systems, Man
  and Cybernetics, 2002 IEEE International Conference on, Vol.~3, 2002, p. 6
  pp. vol.3.

\bibitem{Opp2004}
W.~Opp, F.~Sahin, An artificial immune system approach to mobile sensor
  networks and mine detection, in: Systems, Man and Cybernetics, 2004 IEEE
  International Conference on, Vol.~1, 2004, pp. 947 --952 vol.1.

\bibitem{Greensmith2008}
J.~Greensmith, U.~Aickelin, The deterministic dendritic cell algorithm,
  Artificial Immune Systems (2008) 291--302.

\bibitem{Greensmith2009}
J.~Greensmith, U.~Aickelin, Artificial dendritic cells: Multi-faceted
  perspectives, in: A.~Bargiela, W.~Pedrycz (Eds.), Human-Centric Information
  Processing Through Granular Modelling, Vol. 182 of Studies in Computational
  Intelligence, Springer Berlin / Heidelberg, 2009, pp. 375--395.

\bibitem{Oates2007}
R.~Oates, J.~Greensmith, U.~Aickelin, J.~Garibaldi, G.~Kendall, The application
  of a dendritic cell algorithm to a robotic classifier, Artificial Immune
  Systems (2007) 204--215.

\bibitem{Prieto2008}
C.~Prieto, F.~Nino, G.~Quintana, A goalkeeper strategy in robot soccer based on
  danger theory, in: Evolutionary Computation, 2008. CEC 2008. (IEEE World
  Congress on Computational Intelligence). IEEE Congress on, 2008, pp. 3443
  --3447.

\bibitem{Xiong2007}
G.~Xiong, J.~Gong, H.~Chen, Z.~Su, Multi-robot exploration based on market
  approach and immune optimizing strategy, in: Autonomic and Autonomous
  Systems, 2007. ICAS07. Third International Conference on, 2007, p.~29.

\bibitem{Yuan2009}
M.~Yuan, S.-a. Wang, C.~Wu, K.~Li, Apf-guided adaptive immune network algorithm
  for robot path planning, Frontiers of Computer Science in China 3 (2009)
  247--255, 10.1007/s11704-009-0015-5.

\bibitem{Khatib1986}
O.~Khatib, Real-time obstacle avoidance for manipulators and mobile robots, The
  International Journal of Robotics Research 5~(1) (1986) 90--98.

\bibitem{Cutello2007}
V.~Cutello, M.~Romeo, On the convergence of immune algorithms, in: In
  Proceedings of the 1st IEEE Symposium on Foundations of Computational
  Intelligence, 2007.

\bibitem{Cutello2010}
V.~Cutello, G.~Nicosia, M.~Pavone, G.~Stracquadanio, An information-theoretic
  approach for clonal selection algorithms, in: E.~Hart, C.~McEwan, J.~Timmis,
  A.~Hone (Eds.), Artificial Immune Systems, Vol. 6209 of Lecture Notes in
  Computer Science, Springer Berlin / Heidelberg, 2010, pp. 144--157.

\bibitem{Timmis2008}
J.~Timmis, A.~Hone, T.~Stibor, E.~Clark, Theoretical advances in artificial
  immune systems, Theoretical Computer Science 403~(1) (2008) 11 -- 32.

\bibitem{Aazahin2005}
E.~Sahin, Swarm robotics: From sources of inspiration to domains of
  application, in: E.~Sahin, W.~Spears (Eds.), Swarm Robotics, Vol. 3342 of
  Lecture Notes in Computer Science, Springer Berlin / Heidelberg, 2005, pp.
  10--20.

\bibitem{BRICS2010}
W.~Nowak, A.~Zakharov, S.~Blumenthal, E.~Prassler, Benchmarks for mobile
  manipulation and robust obstacle avoidance and navigation, Tech. rep., Best
  Practice in Robotics (BRICS) (April 2010).

\bibitem{Hui2008}
N.~Hui, D.~Pratihar, Soft computing-based navigation schemes for a real wheeled
  robot moving among static obstacles, Journal of Intelligent and Robotic
  Systems 51~(3) (2008) 333--368.

\bibitem{Mabrouk2008}
M.~H. Mabrouk, C.~R. McInnes, Solving the potential field local minimum problem
  using internal agent states, Robot. Auton. Syst. 56~(12) (2008) 1050--1060.

\bibitem{Barraquand1992}
J.~Barraquand, B.~Langlois, J.-C. Latombe, Numerical potential field techniques
  for robot path planning, Systems, Man and Cybernetics, IEEE Transactions on
  22~(2) (1992) 224 --241.

\bibitem{Zou2003}
X.-y. Zou, J.~Zhu, Virtual local target method for avoiding local minimum in
  potential field based robot navigation, Journal of Zhejiang University -
  Science A 4~(3) (2003) 264--269.

\bibitem{Borenstein1989}
J.~Borenstein, Y.~Koren, Real-time obstacle avoidance for fact mobile robots,
  Systems, Man and Cybernetics, IEEE Transactions on 19~(5) (1989) 1179 --1187.

\bibitem{Borenstein1991}
J.~Borenstein, Y.~Koren, The vector field histogram-fast obstacle avoidance for
  mobile robots, Robotics and Automation, IEEE Transactions on 7~(3) (1991) 278
  --288.

\end{thebibliography}







\end{document}